# Scaling up Heuristic Planning with Relational Decision Trees

**Tomás de la Rosa**　　　　　　　　　　　　　　　　　　TROSA@INF.UC3M.ES
**Sergio Jiménez**　　　　　　　　　　　　　　　　　　SJIMENEZ@INF.UC3M.ES
**Raquel Fuentetaja**　　　　　　　　　　　　　　　　RFUENTET@INF.UC3M.ES
**Daniel Borrajo**　　　　　　　　　　　　　　　　　　DBORRAJO@IA.UC3M.ES
*Departamento de Informática*
*Universidad Carlos III de Madrid*
*Av. Universidad 30, Leganés, Madrid, Spain*

## Abstract

Current evaluation functions for heuristic planning are expensive to compute. In numerous planning problems these functions provide good guidance to the solution, so they are worth the expense. However, when evaluation functions are misguiding or when planning problems are large enough, lots of node evaluations must be computed, which severely limits the scalability of heuristic planners. In this paper, we present a novel solution for reducing node evaluations in heuristic planning based on machine learning. Particularly, we define the task of learning search control for heuristic planning as a relational classification task, and we use an off-the-shelf relational classification tool to address this learning task. Our relational classification task captures the preferred action to select in the different planning contexts of a specific planning domain. These planning contexts are defined by the set of helpful actions of the current state, the goals remaining to be achieved, and the static predicates of the planning task. This paper shows two methods for guiding the search of a heuristic planner with the learned classifiers. The first one consists of using the resulting classifier as an action policy. The second one consists of applying the classifier to generate lookahead states within a Best First Search algorithm. Experiments over a variety of domains reveal that our heuristic planner using the learned classifiers solves larger problems than state-of-the-art planners.

## 1. Introduction

During the last few years, state-space heuristic search planning has achieved significant results and has become one of the most popular paradigms for automated planning. However, heuristic search planners suffer from strong scalability limitations. Even well-studied domains like *Blocksworld* become challenging for these planners when the number of blocks is relatively large. Usually, state-space heuristic search planners are based on action grounding, which makes the state-space to be explored very large when the number of objects and/or action parameters is large enough. Moreover, domain-independent heuristics are expensive to compute. In domains where these heuristics are more misleading, heuristic planners spend most of their planning time computing useless node evaluations. Even with the best current domain-independent heuristic functions in the literature, forward chaining heuristic planners currently have to visit too many nodes, which takes considerable time, especially due to the time required to compute those heuristic functions.

These problems entail strong limitations on the application of heuristic planners to real problems. For instance, logistics applications need to handle hundreds of objects together with hundreds of vehicles and locations (Flórez, García, Torralba, Linares, García-Olaya, & Borrajo, 2010). Current heuristic search planners exhaust the computational resources before solving a problem in a real logistics application.

A classic approach for dealing with planning scalability issues is assisting the search engines of planners with Domain-specific Control Knowledge (DCK). Examples of planning systems that benefit from this knowledge are TLPLAN (Bacchus & Kabanza, 2000), TALPLANNER (Doherty & Kvarnström, 2001) and SHOP2 (Nau, Au, Ilghami, Kuter, Murdock, Wu, & Yaman, 2003). Nevertheless, hand-coding DCK is a complex task because it implies expertise in both, the planning domain and the search algorithm of the planning system. In recent years there has been a renewed interest in using Machine Learning (ML) to automatically extract DCK. Zimmerman and Kambhampati (2003) made a comprehensive survey of ML for defining DCK. As shown in the first *learning for planning* competition held in 2008 (Learning Track), this renewed interest is specially targeted at heuristic planners.

This paper presents an approach for learning DCK for planning by building domain-dependent relational decision trees from examples of good quality solutions of a forward-chaining heuristic planner. These decision trees are built with an off-the-shelf relational classification tool and capture which is the best action to take for each possible decision of the planner in a given domain. The resulting decision trees can be used either as a policy to solve planning problems directly or to generate lookahead states within a Best First Search (BFS) algorithm. Both techniques allow the planner to avoid state evaluations, which helps in the objective of improving scalability. The approach has been implemented in a system we have called ROLLER. This work is an improvement of a previous one (De la Rosa, Jiménez, & Borrajo, 2008). Alternatively, a ROLLER version for repairing relaxed plans (De la Rosa, Jiménez, García-Durán, Fernández, García-Olaya, & Borrajo, 2009) competed in the Learning Track of the 6th International Planning Competition (IPC) held in 2008. ROLLER improvements presented in this article are mainly a result of lessons learned from the competition, which will be discussed later.

The paper is organized as follows. Section 2 introduces the issues that need to be considered when designing a learning system for heuristic planning. They will help us to clarify which decisions we made in the development of our approach. Section 3 describes the ROLLER system in detail. Section 4 presents the experimental results obtained in a variety of benchmarks. Section 5 discusses the improvements of the ROLLER system compared to the previous version of the system. Section 6 revises the related work on learning DCK for heuristic planning. Finally, the last section discusses some conclusions and future work.

## 2. Common Issues in Learning Domain-specific Control Knowledge

When designing an ML process for the automatic acquisition of DCK, one must consider some common issues, among others:

1. **The representation of the learned knowledge**. Predicate logic is a common language to represent planning DCK because planning tasks are usually defined in this language. However, other representation languages have been used aiming to make the learning of DCK more effective. For instance, languages for describing object classes such as the *Concept Language* (Martin & Geffner, 2000) or *Taxonomic Syntax* (Mcallester & Givan, 1989) have been shown to provide a useful learning bias for different domains.

   Another representation issue is the selection of the feature space (i.e., the set of instance features used for representing the learned knowledge and for training the system.). The feature space should be able to capture the *key* knowledge of the domain. Traditionally, the feature space consisted only of predicates for describing the current state and the goals of the planning task. The feature space can be enriched with extra predicates, called *metapredicates*, which capture extra useful information of the planning context such as the applicable operators or the pending goals (Veloso, Carbonell, Pérez, Borrajo, Fink, & Blythe, 1995). Recently, works on learning DCK for heuristic planners define *metapredicates* to capture information about the planning context of a heuristic planner, including for example, predicates which capture the actions in the relaxed plan of a given state (Yoon, Fern, & Givan, 2008).

2. **The learning algorithms**. Inductive Logic Programming (ILP) (Muggleton & De Raedt, 1994) deals with the development of inductive techniques which learn a given target concept from examples described in predicate logic. Because planning tasks are normally represented in predicate logic, ILP algorithms are quite suitable for DCK learning. Moreover, in recent years, ILP has broadened its scope to cover the whole spectrum of ML tasks such as regression, clustering and association analysis, extending the classical propositional ML algorithms to the relational framework. Consequently, ILP algorithms have been used by heuristic planners to capture DCK in different forms such as decision rules to select actions in the different planning context or regression rules to obtain better node evaluations (Yoon et al., 2008).

3. **The generation of training examples**. The success of ML algorithms depends directly on the quality of the training examples used. When learning planning DCK, these examples are extracted from the experience collected from solving training problems, which should be representative of different tasks across the domain. Therefore, the quality of the training examples will depend on the variety of the problems used for training and the quality of the solutions to these problems. Traditionally, these training problems are obtained by random generators provided with some parameters to tune problems difficulty. In this way, one has to find, for each domain, which kind of problems makes the learning algorithm generalize useful DCK.

4. **Use of the learned DCK**. Decisions made for each of these three issues affect the quality of the learned DCK. Some representation schemes may not be expressive enough to capture effective DCK for a given domain, the learning algorithm may not be able to acquire the useful DCK within reasonable time and memory requirements, or the set of training problems may lack significant examples of the *key* knowledge. In all these situations, a direct use of the learned DCK will not improve the scalability of the planner, and could even decrease its performance. An effective way of dealing with this problem in heuristic planners is integrating the learned DCK within robust strategies such as a Best-First Search (Yoon et al., 2008) or combining it with domain-independent heuristic functions (Röger & Helmert, 2010).

## 3. The ROLLER System

This section describes how the general scheme for learning DCK is instantiated in the ROLLER system. First, it describes the DCK representation followed by ROLLER. Second, it explains the learning algorithm used by ROLLER. Third, it depicts how ROLLER collects good quality training examples and finally, it shows different approaches for scaling up heuristic planning algorithms with the learned DCK.

### 3.1 The Representation of the Learned Knowledge: *Helpful Contexts* in Heuristic Planning

We present our approach following the notation specified by the Planning Domain Definition Language (PDDL) for typed STRIPS tasks. Accordingly, the definition of a planning domain $\mathcal{D}$ comprises the definition of:

- A hierarchy of types.

- A set of typed constants, $\mathcal{C}_{\mathcal{D}}$, representing the objects present in all tasks for the domain. This set can be empty.

- A set of predicate symbols, $\mathcal{P}$, each one with its corresponding arity and the type of its arguments.

- A set of operators $\mathcal{O}$, whose arguments are typed variables.

Variables are declared directly when defining each operator argument, so they are local to the operator definition. We will call $\mathcal{P}_o$ the set of atomic formulas that can be generated using the defined predicates $\mathcal{P}$, the variables defined as arguments of the operator $o \in \mathcal{O}$, and the general constants $\mathcal{C}_{\mathcal{D}}$. Then, each operator $o \in \mathcal{O}$ is defined by three sets: $pre(o) \subseteq \mathcal{P}_o$, the operator preconditions; $add(o) \subseteq \mathcal{P}_o$, the positive effects; and $del(o) \subseteq \mathcal{P}_o$, the negative effects of the operator.

A planning task $\Pi$ for the domain $\mathcal{D}$ is a tuple $< \mathcal{C}_\Pi, s_0, G >$ where $\mathcal{C}_\Pi$ is a set of typed constants representing the objects which are particular to the task, $s_0$ is the set of ground atomic formulas describing the initial state and $G$ is the set of ground atomic formulas describing the goals. Given the total set of constants $\mathcal{C} = \mathcal{C}_\Pi \cup \mathcal{C}_{\mathcal{D}}$, the task $\Pi$ defines a finite state space $\mathcal{S}$ and a finite set $\mathcal{A}$ of instantiated operators over $\mathcal{O}$. A state $s \in \mathcal{S}$ is a set of ground atomic formulas representing the facts that are true in $s$. States are described following the closed world assumption. An instantiated operator or action $a \in \mathcal{A}$ is an operator where each variable has been replaced by a constant in $\mathcal{C}$ of the same type. Thus, $\mathcal{A}$ is the set of all actions $a$ that can be generated using the set of constants $\mathcal{C}$ and the set of operators $\mathcal{O}$. Under this definition, solving a planning task $\Pi$ implies finding a plan $\pi$ as the sequence of actions $(a_1, \ldots, a_n), a_i \in \mathcal{A}$ that transforms the initial state into a state in which the goals are achieved.

The planning contexts defined by ROLLER rely on the concepts of *relaxed plan heuristic* and *helpful actions*, both introduced by the FF planner (Hoffmann & Nebel, 2001). The relaxed plan heuristic returns an integer for each evaluated node, which is the number of actions in a solution to the relaxed planning task $\Pi^+$ from that node. $\Pi^+$ is a simplification of the original task in which the deletes of actions are ignored. The idea of delete-relaxation for computing heuristics in planning was first introduced by McDermott (1996) and by Bonet, Loerincs and Geffner (1997).

The relaxed plan is extracted from a relaxed planning graph, which is a sequence of facts and actions layers $(F_0, A_0, \ldots, A_t, F_t)$. The first fact layer contains all facts in the initial state. Then each action layer contains the set of all applicable actions given the previous fact layer. Each fact layer contains the set of all positive effects of all actions appearing in the previous layers. The process finishes when all the goals are in a fact layer, or when two consecutive facts layers have the same facts. In the last case, the relaxed problems have no solution and the relaxed plan heuristic returns infinity.

Once the relaxed planning graph is built, the solution is extracted in a backwards process. Each goal appearing for the first time in fact layer $i$ is assigned to the set of goals of that layer, $G_i$. Then, from the last set of goals, $G_t$, to the second set of goals, $G_1$, and for each goal in each goals set, an action is selected which generates the goal and whose layer index is minimal. Afterwards, each precondition of that action (i.e. a subgoal) is included in the goals set corresponding to the first layer where this fact appears. When the process is finished, the set of selected actions comprises the relaxed plan.

According to the extraction process, the FF planner marks as helpful actions the set of actions in the first layer $A_0$ of the relaxed planning graph which can achieve any of the subgoals of the next fact layer, i.e. in the goals set $G_1$. In other words, helpful actions are those applicable actions which generate facts that are top-level goals of the problems or required by any action of the relaxed plan. Formally, the set of helpful actions of a given state $s$ is defined as:

$$helpful(s) = \{a \in A_0 \mid add(a) \cap G_1 \neq \emptyset\}$$

The FF planner uses helpful actions in the search as a pruning technique, because they are considered as the only candidates for being selected during the search. Given that each state generates its own particular set of helpful actions, we claim that the helpful actions, together with the remaining goals and the static literals of the planning task, encode a *helpful context* related to each state. The *helpful actions* and the remaining *target goals* relate actions that are more likely to be applied with the goals that need to be achieved. These relations arise because helpful actions and target goals often share some arguments (problem objects). Additionally, the *static predicates* express facts that characterize objects of the planning task. Identifying these objects is also relevant since they may be shared arguments for helpful actions and/or target goals.

**Definition 1** *The* **helpful context** *for a state $s$ is defined as*

$$\mathcal{H}(s) = \{helpful(s), target(s), static(s)\}$$

*where $target(s) \subseteq G$ describes the set of goals not achieved in the state $s$, $target(s) = G - s$ and $static(s)$ is the set of literals that always hold in the planning task. They are defined in the initial state and are present at every state given that they can not be changed by any action. Thus, $static(s) = \{p \in s \mid \nexists a \in A : p \in add(a) \vee p \in del(a)\}$.*

The helpful context is an alternative representation to the tuple $<state, goals, applied\_action>$, traditionally used when learning DCK for planning. Helpful contexts present some advantages for improving the scalability of heuristic planners:

- In most domains, the set of helpful actions contains the actions most likely to be applied and focusing reasoning on them has been shown to be a good strategy.

- The set of helpful actions is normally smaller than the set of non-static literals in the state (i.e., $s - static(s)$). Thus, the process of matching learned DCK within the search obtains the benefits of using a more compact representation.

- The number of helpful actions normally decreases when the search has fewer goals left. Therefore, the matching process will become faster when the search is advancing towards the goals.

## 3.2 The Learning Algorithm: Learning Generalized Policies with Relational Decision Trees

ROLLER implements a two-step learning process for building DCK from a collection of examples from different helpful contexts:

1. Learning the operator classifier. ROLLER builds a classifier to choose the best operator in the different helpful contexts.

2. Learning the binding classifiers. For each operator in the domain, ROLLER builds a classifier to choose the best binding (instantiation of the operator) in the different helpful contexts.

The learning process is split into these two steps to build DCK with off-the-shelf learning tools. Each planning action may have different number of arguments and arguments of different types (e.g. actions `switch_on(instrument, satellite)` and `turn_to(satellite, direction, direction)` from the *Satellite* domain) which hinders the definition of the target classes. This two-step decision process is also clearer from the decision-making point of view. It helps users to understand the generated DCK better by focusing on either the decision of which operator to apply or which bindings to use for a given selected operator. Both the learning algorithm and the set of learning examples are the same for the two learning steps. Figure 1 shows an overview of the learning process of the ROLLER system.

## Roller Learner

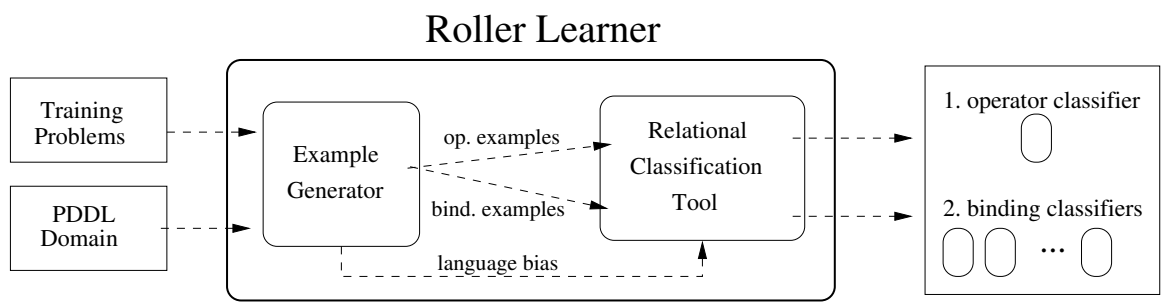

Figure 1: Overview of the ROLLER learning process.

### 3.2.1 LEARNING RELATIONAL DECISION TRESS

A classic approach to assist decision making consists of gathering a significant set of previous decisions and building a decision tree that generalizes them. The leaves of the resulting tree contain the classes (decisions to make), and the internal nodes contain the conditions that lead to those decisions. The most common way to build these trees is following the *Top-Down Induction of Decision Trees* (TDIDT) algorithm (Quinlan, 1986). This algorithm builds the tree by repeatedly splitting the set of training examples according to the conditions that minimize the entropy in the examples. Traditionally, training examples are described in an attribute-value representation. Therefore, conditions of the decision trees represent tests over the value of a given attribute of the examples. Nevertheless, this attribute-value approach is not suitable for representing decisions if we want to keep the predicate logic representation. A better approach is to represent decisions relationally, for instance, a given action is chosen to reach certain goals in a given context if they share some arguments. Recently, new algorithms for building relational decision trees from examples described as predicate logic facts have been developed. These new relational learning algorithms are similar to the propositional ones, except that (1) condition nodes in the tree do not refer to attribute values, but to logic queries about relational facts holding in the training examples and (2), these logic queries can share variables with condition nodes placed above in the decision tree. The learning algorithm is a greedy search process. Since the space of potential relational decision trees is usually huge, this search is normally biased according to a specification of syntactic restrictions called *language bias*. This specification contains the target concept, the predicates that can appear in the condition nodes of the trees and some learning-specific knowledge such as type information, or input and output variables of predicates.

In this paper we use the tool TILDE (Blockeel & De Raedt, 1998) for learning the operator and binding classifiers. This tool implements a relational version of the TDIDT algorithm, although any other off-the-shelf tool for learning relational classifiers could have been used, such as PRO-GOL (Muggleton, 1995) or RIBL (Emde & Wettschereck, 1996). Each of these different learning algorithms would provide different results, since they explore the classifiers space differently. The study of the pros and cons of the different algorithms is beyond the scope of the paper. For a comprehensive explanation of current relational learning approaches please refer to the work by De Raedt (2008).

### 3.2.2 LEARNING THE OPERATOR CLASSIFIER

The inputs to learning the operator classifier are:

- **Training examples**. Examples are represented in a Prolog-like syntax and consist of the operator selected (the class) together with the helpful context (the background knowledge in terms of relational learning) in which it was selected. In particular, an example contains:

  - *Class*. We use the predicate of arity 3 `selected` to encode the operator chosen in the context. This predicate is the target concept of this learning step. Its first argument holds the example identifier that links the rest of the example predicates. The second argument is the problem identifier, which links the *static predicates* shared by all examples coming from the same planning problem. The third argument is the example class, i.e., the name of the selected operator in the helpful context.

  - *Helpful predicates*. They are predicates to express the *helpful actions* contained in the helpful context. The predicate symbol of these predicates is `helpful_`$a_i$ where $a_i$ is the name of an instantiated action. The arguments are the example and problem identifier together with the parameters of action $a_i$. As it is an instantiated action, its parameters are constants.

  - *Target goal predicates*. They represent the predicates that appear in the goals and do not hold in the current state. These predicates have the form `target_goal_`$g_i$ where $g_i$ are the domain predicates. Each predicate also contains the example and problem identifiers.

  - *Static predicates*. They represent the static predicates of a given problem. These predicates are shared by all the training examples that belong to the same planning problem. They have the form `static_fact_`$f_i$ where $f_i$ are the domain predicates that do not appear in the effects of any domain action. They have as arguments the problem identifier and the corresponding arguments of each domain predicate.

Figure 2 shows one learning example with id `tr01_e1` consisting of a selection of the operator `switch-on` and its associated helpful context. This example is used for building the operator classifier for the *Satellite* domain.

```
% Example tr01_e1 from problem tr01
selected(tr01_e1,tr01,switch_on).
helpful_turn_to(tr01_e1,tr01,satellite0,groundstation1,star0).
helpful_turn_to(tr01_e1,tr01,satellite0,phenomenon2,star0).
helpful_turn_to(tr01_e1,tr01,satellite0,phenomenon3,star0).
helpful_turn_to(tr01_e1,tr01,satellite0,phenomenon4,star0).
helpful_switch_on(tr01_e1,tr01,instrument0,satellite0).
target_goal_have_image(tr01_e1,tr01,phenomenon3,infrared2).
target_goal_have_image(tr01_e1,tr01,phenomenon4,infrared2).
target_goal_have_image(tr01_e1,tr01,phenomenon2,spectrograph1).

% Static Predicates of problem
static_fact_calibration_target(tr01,instrument0,groundstation1).
static_fact_supports(tr01,instrument0,spectrograph1).
static_fact_supports(tr01,instrument0,infrared2).
static_fact_on_board(tr01,instrument0,satellite0).
```

Figure 2: Knowledge base corresponding to an example from the Satellite domain. The example has the id `tr01_e1`, which links all example predicates. It was obtained solving the training problem `tr01` which links the rest of examples for the same problem. The selected operator in this helpful context is **switch_on**, which corresponds to one of the helpful actions encoded in the helpful predicates of the example.

- **Language bias**: This bias specifies constraints over the arguments of the predicates in the training examples. We do not assume any domain-specific constraint, given that our learning technique is domain-independent. So, this bias only contains restrictions over argument types and restrictions which ensure that identifier variables are not added as new variables in the classifier generation. This bias is automatically extracted from the PDDL domain definitions and consists of a declaration of the predicates used in the learning example and their argument types. Figure 3 shows the language bias specified for learning the operator classifier for the *Satellite* domain.

The resulting relational decision tree represents a set of disjoint rules for action selection that can be used to provide advice to the planner: *the internal nodes* of the tree contain the set of conditions related to the helpful context under which the advice can be provided. The *leaf nodes* contain the corresponding advice; in this case, the operator to select and the number of examples covered by the rule. The operator to select is the one which has been selected more often in the training examples covered by the rule. The operator classifiers learned by ROLLER also advise on *non-helpful actions*. Given a state, non-helpful actions are the subset of applicable actions in the state that are not considered as helpful actions. Certainly, these actions are not part of the helpful contexts defined. However, the learned operator classifiers indicate the name of the operator to select regardless of whether it was helpful or not. Figure 4 shows the operator tree learned for the Satellite

```
% ---- The target concept ----
predict(selected(+IdExample,+IdProblem,-Operator)).
type(selected(idex,idprob,class)).
classes([turn_to,switch_on,switch_off,calibrate,take_image]).

% ---- The helpful context ----
% predicates for the helpful actions
rmode(helpful_turn_to(+IdExample,+IdProblem,+-S1,+-D1,+-D2)).
type(helpful_turn_to(idex,idprob,satellite,direction,direction)).

rmode(helpful_switch_on(+IdExample,+IdProblem,+-I1,+-S1)).
type(helpful_switch_on(idex,idprob,instrument,satellite)).

rmode(helpful_switch_off(+IdExample,+IdProblem,+-I1,+-S1)).
type(helpful_switch_off(idex,idprob,instrument,satellite)).

rmode(helpful_calibrate(+IdExample,+IdProblem,+-S1,+-I1,+-D1)).
type(helpful_calibrate(idex,idprob,satellite,instrument,direction)).

rmode(helpful_take_image(+IdExample,+IdProblem,+-S1,+-D1,+-I1,+-M1)).
type(helpful_take_image(idex,idprob,satellite,direction,instrument,mode)).

% predicates for the target goals
rmode(target_goal_pointing(+IdExample,+IdProblem,+-S1,+-D1)).
type(target_goal_pointing(idex,idprob,satellite,direction)).

rmode(target_goal_have_image(+IdExample,+IdProblem,+-D1,+-M1)).
type(target_goal_have_image(idex,idprob,direction,mode)).

% predicates for the static facts
rmode(static_fact_on_board(+IdProblem,+-I1,+-S1)).
type(static_fact_on_board(idprob,instrument,satellite)).

rmode(static_fact_supports(+IdProblem,+-I1,+-M1)).
type(static_fact_supports(idprob,instrument,mode)).

rmode(static_fact_calibration_target(+IdProblem,+-I1,+-D1)).
type(static_fact_calibration_target(idprob,instrument,direction)).
```

Figure 3: Language bias for learning the operator classifier of the *Satellite* domain. It is automatically generated from the PDDL definition. `rmode` predicates indicate those which can be used in the tree. `type` predicates indicate types for each particular `rmode`.

domain. In learned decision trees each branch is denoted by the symbols `+--:<yes/no>`, where `yes` indicates the next node for positive answers to the current question and `no` indicates the next node for negative answers. In the figure, the first branch states that when there is a `calibrate` action in the set of helpful actions, the recommendation (in square brackets) is choosing that action (i.e. `calibrate`). In addition, the branch indicates that the recommended action has occurred 44 times in the training examples. Moreover, each leaf node has information (in double square brack-ets) about the number of times each type of action has been selected in the training examples covered by the rule in the current branch. Thus, in our case, the action `calibrate` has been selected 44 out of a total of 44 times, and other operators have never been selected. The second branch says that if there is no `calibrate` helpful action, but there is a `take_image` one, the planner selected to `take_image` 110 out of 110 times. If there are no helpful `calibrate` or `take_image` actions but there is a helpful `switch_on` action, `switch_on` is the recommendation, that has been selected 44 out of 59 times. Other tree branches are interpreted similarly.

```
selected(-A,-B,-C)
helpful_calibrate(A,B,-D,-E,-F) ?
+--yes:[calibrate] 44.0 [[turn_to:0.0,switch_on:0.0,switch_off:0.0,
|                         calibrate:44.0,take_image:0.0]]
+--no:  helpful_take_image(A,B,-G,-H,-I,-J) ?
        +--yes:[take_image] 110.0 [[turn_to:0.0,switch_on:0.0,switch_off:0.0,
        |                            calibrate:0.0,take_image:110.0]]
        +--no:  helpful_switch_on(A,B,-K,-L) ?
                +--yes:[switch_on] 59.0 [[turn_to:15.0,switch_on:44.0,
                |                           switch_off:0.0,calibrate:0.0,
                |                           take_image:0.0]]
                +--no:  [turn_to] 149.0 [[turn_to:149.0,switch_on:0.0,
                                           switch_off:0.0,calibrate:0.0,
                                           take_image:0.0]]
```

Figure 4: Relational decision tree learned for the operator selection in the *Satellite* domain. Internal nodes (with "?" ending) have queries to helpful contexts. Leaf nodes (in brackets) have the class and the number of observed examples for each operator.

### 3.2.3 LEARNING THE BINDING CLASSIFIERS

At the second learning step, a relational decision tree is built for each domain operator $o \in \mathcal{O}$. These trees indicate the bindings to select for $o$ in the different helpful contexts. The inputs for learning the binding classifier of operator $o$ are:

- **Training examples**. These consist exclusively of the helpful contexts where operator $o$ was selected, together with the applicable instantiations of $o$ in these contexts. Note that for a given helpful context, the applicable instantiations of $o$ may include both helpful and non-helpful actions. Helpful contexts are coded exactly as in the previous learning step. The applicable instantiations of $o$ are represented with the `selected_o` predicate. This predicate is the target concept of the second learning step and its arguments are the example and problem identifiers, the instantiated arguments of the applicable action and the example class (*selected* or *rejected*). The purpose of this predicate is to distinguish between good and bad bindings for the operator. Figure 5 shows a piece of the knowledge base for building the binding tree corresponding to the action `switch_on` from the *Satellite* domain. This example, with id `tr07_e63`, resulted in the selection of the action instantiation

switch_on(instrument1,satellite0). The action switch_on(instrument0, satellite0) was also applicable but it was *rejected* by the planner.

- **Language bias**: The bias for learning binding trees is the same as the bias for learning the operator tree, except that it includes the definition of the selected_o predicate. As in the previous learning step, the *language bias* for learning a binding tree is also automatically extracted from the PDDL domain definition. Figure 6 shows part of the language bias specified for learning the binding tree for the action switch_on from the *Satellite* domain.

```
% Example tr07_e63 from problem tr07
selected_switch_on(tr07_e63,tr07,instrument0,satellite0,rejected).
selected_switch_on(tr07_e63,tr07,instrument1,satellite0,selected).
helpful_switch_on(tr07_e63,tr07,instrument0,satellite0).
helpful_switch_on(tr07_e63,tr07,instrument1,satellite0).
helpful_turn_to(tr07_e63,tr07,satellite0,star1,star2).
helpful_turn_to(tr07_e63,tr07,satellite0,star5,star2).
helpful_turn_to(tr07_e63,tr07,satellite0,phenomenon7,star2).
helpful_turn_to(tr07_e63,tr07,satellite0,phenomenon8,star2).
target_goal_have_image(tr07_e63,tr07,phenomenon8,spectrograph2).
target_goal_have_image(tr07_e63,tr07,phenomenon7,spectrograph2).
target_goal_have_image(tr07_e63,tr07,star5,image1).

% Static Predicates of problem
static_fact_calibration_target(tr07,instrument0,star1).
static_fact_calibration_target(tr07,instrument1,star1).
static_fact_supports(tr07,instrument0,image1).
static_fact_supports(tr07,instrument1,spectrograph2).
static_fact_supports(tr07,instrument1,image1).
static_fact_supports(tr07,instrument1,image4).
static_fact_on_board(tr07,instrument0,satellite0).
static_fact_on_board(tr07,instrument1,satellite0).
```

Figure 5: Knowledge base corresponding to the example tr07_e63 obtained by solving the training problem tr07 from the *Satellite* domain.

The result of this second learning step is a relational decision tree $t_o$ for each uninstantiated operator $o \in \mathcal{O}$. $t_o$ consists of the set of disjoint rules for the binding selection of $o$. Figure 7 shows an example of the binding tree $t_{switch\_on}$ built for operator switch_on from the *Satellite* domain. According to this tree, the first branch states that when there is a helpful action which is a switch_on of instrument $C$ in satellite $D$, these switch_on bindings $(C, D)$ were selected by the planner 213 out of 249 times. Note that binding trees learned by ROLLER also advise on non-helpful actions. Frequently, the $selected\_o$ predicate matches with tree queries that refer to $helpful\_o$ predicates. In these cases, the no-branch of the query may cover bindings of non-helpful actions for this operator.

For the other binding trees of the *Satellite* domain we refer the reader to the Online Appendix of this article, where we include the learned DCK for the domains used in the experimental section.

```
% ---- The target concept ----
predict(selected_switch_on(+IdExample,+IdProblem,+INST0,+SAT1,-Class)).
type(selected_switch_on(idex,idprob,instrument,satellite,class)).
classes([selected,rejected]).

% ---- The helpful context ----, the same as in the operator classification
...
```

Figure 6: Part of the language bias for learning the binding tree for the `switch_on` action from the *Satellite* domain.

```
selected_switch_on(-A,-B,-C,-D,-E)
helpful_switch_on(A,B,C,D) ?
+--yes:  [selected] 249.0 [[selected:213.0,rejected:36.0]]
+--no:   [rejected] 63.0 [[selected:2.0,rejected:61.0]]
```

Figure 7: Relational decision tree learned for the bindings selection of the `switch_on` action from the *Satellite* domain.

In many cases, decision trees are somewhat more complex that the one shown in Figure 7. For instance, the *turn_to* binding tree has 29 nodes and includes several queries about target goals (e.g., asking if there is a pending image at the new pointed direction) and others about static facts (e.g., asking if the new pointed direction is a calibration target).

### 3.3 Generation of Training Examples

ROLLER training examples are instances of decisions made when solving training problems. In order to characterize a variety of good solutions, these decisions should consider different alternatives for solving each individual problem. At a given search tree node (state), the alternatives come from the possibility of choosing different operators or of having different bindings for a single operator, in both cases assuming the alternative will lead to equally good solutions.

Regarding binding decisions, if actions from some alternative solutions are ignored, they are tagged as *rejected* and consequently they introduce noise in the learning process. For instance, consider the problem of Figure 8 from the *Satellite* domain in which a satellite, with a calibrated instrument, must turn to directions $D1, D2$ and $D3$ in order to take images there. In this planning context, the three `turn_to` actions are helpful actions and regarding only one solution makes learning consider one action as *selected* and the other two actions as *rejected*. However, the learned knowledge should always recommend a helpful `turn_to` action towards a direction where the satellite (with the corresponding calibrated instrument) has to take an image. To learn such kind of knowledge, ROLLER should consider the three `turn_to` actions as *selected* because the three actions correspond to selectable actions for learning the correct knowledge in this particular planning context. If most of these actions are marked as *rejected* the learner will consider selecting `turn_to` in the described context as a bad choice.

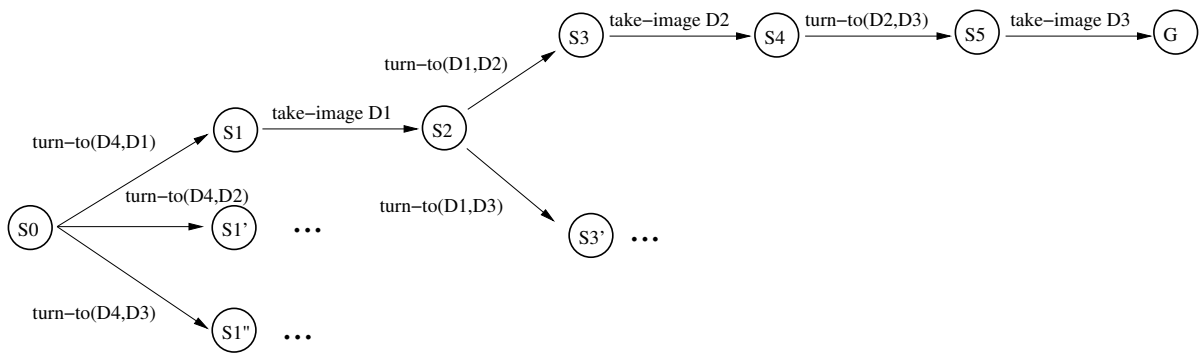

Figure 8: Solution path alternatives in a simplified Satellite problem.

Regarding operator decisions, complete training with a full catalogue of different solutions can confuse the learning process. For instance, consider the example problem of Figure 9 where the goal is to take an image at direction $D2$. Before applying `calibrate` action in $s_2$, it is necessary to switch on the instrument $T$ and to turn the satellite to $D1$ (the calibration target direction). These two actions are helpful in $s_o$ and generate two different solution paths. In fact, they are commutative. Generalizing operator selection from these kinds of helpful contexts is difficult when the training examples contain examples of both types (i.e. examples where the `switch-on` action is situated before the `turn-to` action and vice versa). This is caused by the fact that for the same helpful context there are different operators to choose from and all of them are equally good choices.

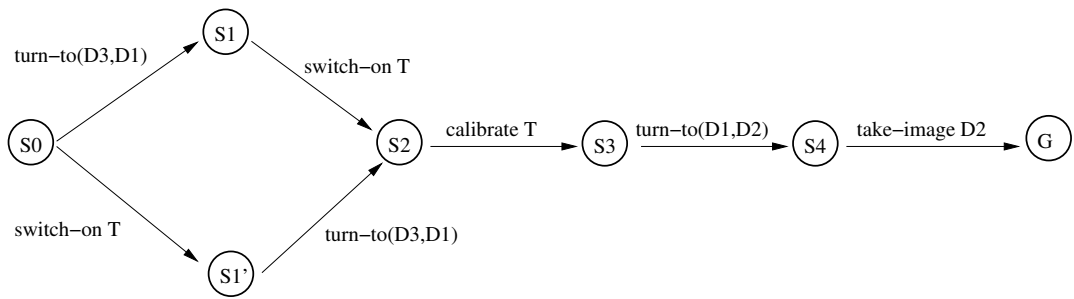

Figure 9: Solution path alternatives in a simplified Satellite problem.

ROLLER follows a commitment approach for the generation of training examples: (1) **Generation of solutions**. Given a training problem, ROLLER performs an exhaustive search to obtain multiple best-cost solutions, taking into account the alternatives of different binding choices. (2) **Selection of solutions**. ROLLER selects a subset of solutions from the set of best-cost solutions in order to reproduce a particular preference for the operator alternatives. (3) **Extraction of examples from solutions**. ROLLER encodes the selected subset of solutions as examples for the required learning, operator classification or binding classification. The following sections detail how ROLLER proceeds at each of these three steps.

### 3.3.1 GENERATION OF SOLUTIONS

ROLLER solves each training problem using a *Best-First Branch and Bound* (BFS-BnB) algorithm that extracts multiple good-quality solutions. If the search space has not been explored exhaustively within a time bound, the problem is discarded and no examples are generated from it. Therefore, training problems need to be sufficiently small. In addition, training problems need to be representative enough to generalize the DCK which assists ROLLER when solving future problems in the same domain.

The BFS-BnB search is completed without pruning repeated states. In practice, many repeated states are generated by changing the order among actions of different solution paths. Thus, pruning repeated states would involve tagging actions leading to these solutions as *rejected* bindings, which is in fact not true. In addition, the BFS-BnB algorithm prunes according to the A* evaluation function $f(n) = g(n) + h(n)$, where $g(n)$ is the node cost (in this work we use plan length as the cost function) and $h(n)$ is the FF heuristic. The safe way to prune the search space is by using an admissible heuristic. However, existing admissible heuristics will not allow ROLLER to complete an exhaustive search in problems of reasonable size. In practice, using the FF heuristic produces few overestimations which introduces negligible noise into the learning process. At the end of the search, the BFS-BnB algorithm returns the set of solutions with the best cost. These solutions are used to tag the nodes in the search tree that belong to any of the solutions with the label *on_solution*.

### 3.3.2 SELECTING SOLUTIONS

From the set of best-cost solutions found, ROLLER selects the subset of solutions that will be used for generating training examples. Since it is difficult to develop domain-independent criteria for systematically selecting solutions that reproduce the same operator selection in a particular context, we have defined an approach which, heuristically, prefers some actions over others. These preferences are:

- *Least-commitment preference*: Prefer actions that generate more alternatives of different solution paths.

- *Difficulty preference*: Prefer actions that reach the goals or sub-goals which are most difficult to achieve. In the example of Figure 9 having instrument $T$ switched on is only achievable by one action. On the other hand, pointing to direction $D1$ is considered easier since it can be achieved with actions `turn_to(D2,D1)` and `turn_to(D3,D1)`.

Given $\pi' = a_1, \ldots, a_n$, a best-cost plan for a planning task, we compute these preferences with functions depending on each action.

$$\varphi_{\text{commitment}}(a_i) = |\ \{a' \mid a' \in successors(a_i) \wedge on\_solution(a')\}\ |$$

where the function $successors(a_i)$ returns all applicable actions in state $s_{i+1}$ and function $on\_solution(a)$ verifies whether an action is tagged as being part of a best-cost plan.

$$\varphi_{\text{difficulty}}(a_i) = \frac{1}{\min_{l \in add(a_i)} |\ supporters(l)\ |}$$

where the function $supporters(l) = \{a \in A \mid l \in add(a)\}$ returns the set of actions that achieve the literal $l$.

Solutions are ranked according to these preferences. The ranking for each solution $\pi' = a_1, \ldots, a_n$ is computed as the weighted sum of the action preferences, as follows:

$$ranking(\pi', \varphi) = \sum_{i=0,\ldots,n-1} \frac{(n-i)}{n} \times \varphi(a_{i+1})$$

where $n$ is the plan length and $\varphi$ is one of $\varphi_{\text{commitment}}$ or $\varphi_{\text{difficulty}}$. This sum is weighted to give more importance to the preferences in the first actions of the plan. The first action preference value is multiplied by 1, the second by $(n-1)/n$, and so on. Otherwise, several alternatives (i.e., commutative actions in different positions in a plan) would lead to the same ranking value. We compute the ranking for all best-cost solutions using $\varphi_{\text{commitment}}$. Ties in this ranking are broken with the ranking computed with $\varphi_{\text{difficulty}}$. The subset of solutions with the best ranking values is the subset of solutions selected for generating training examples.

### 3.3.3 EXTRACTING EXAMPLES FROM SOLUTIONS

ROLLER takes the subset of solutions selected at the previous step and generates training examples. When generating examples for the operator classification, ROLLER takes solution plans $\pi' = \{a_1, a_2, \ldots, a_n\}$ which correspond to the sequence of state transitions $\{s_0, s_1, \ldots, s_n\}$ and generates one learning example for each pair $< s_i, a_{i+1} >$ consisting of $\mathcal{H}(s_i)$ and the class (i.e., the operator name of action $a_{i+1}$). See learning example shown in Figure 2.

When generating examples for the binding classification of operator $o$, ROLLER only considers pairs $< s_i, a_{i+1} >$ where $a_{i+1}$ matches operator $o$. A learning example generated from the pair $< s_i, a_{i+1} >$ for the binding selection of the operator $o$ consists of $\mathcal{H}(s_i)$ and the classes of all applicable actions in $s_i$ that match $o$, including $a_{i+1}$. Applicable actions with the *on_solution* label belong to the *selected* class and all other applicable actions to the *rejected* class. Moreover, actions belonging to solutions not in the top ranking are still marked as *selected* even though they are not nodes from which an example is generated. See learning example shown in Figure 5.

## 3.4 Use of the Learned Knowledge: Planning with Relational Decision Trees

This section details how to make heuristic planning benefit from our DCK, beginning with how we build action orderings with the learned DCK. Then, it explains two different search strategies to exploit these orderings: (1) the application of the DCK as a generalized action policy (*Depth-First H-Context Policy* algorithm) and (2) the use of the DCK to generate *lookahead states* within a Best-First Search (BFS) guided by the FF heuristic (*H-Context Policy Lookahead-BFS* algorithm).

### 3.4.1 ORDERING ACTIONS WITH RELATIONAL DECISION TREES

Given a state $s$, the expression $app(s)$ denotes the set of actions applicable in $s$. The learned DCK provides an ordering for $app(s)$. The ordering is built by matching each action $a \in app(s)$ first with the operator classifier and then with the corresponding binding classifier. Figure 10 shows in detail the algorithm for ordering applicable actions with relational decision trees.

The algorithm divides the set of applicable actions into two subsets: the *helpful actions*, and the non-*helpful actions*. Then, it matches the *helpful context* of the state, i.e., $\mathcal{H}(s)$, with the tree for the operator classification. This matching provides a leaf node that contains the list of operators sorted by the number of examples covered by the leaf during the training phase (see the operator

---

**DT-Filter-Sort** *(A,ℋ,T):sorted list of applicable actions*

---

*A*: List of actions
*ℋ*: Helpful Context
*T*: Decision Trees

---

*selected-actions* = ∅
*HA* = helpful-actions(*A*, *ℋ*)
*NON-HA* = *A* \ *HA*
*leaf-node* = classify-operators-tree(*T*, *ℋ*)
**for each** *a* in *HA* **do**
   *priority(a)* = leaf-node-operator-value(*leaf-node*, *a*)
   **if** *priority(a) > 0* **then**
      *(selected(a),rejected(a))* = classify-bindings-tree(*T*, *ℋ*, *a*)
      *selection_ratio(a)*=$\frac{selected(a)}{selected(a)+rejected(a)}$
      *priority(a)* = *priority(a)* + *selection_ratio(a)*
      *selected-actions* = *selected-actions* ∪{*a*}
*max-HA-priority* = $\max_{a \in selected\text{-}actions} priority(a)$
**for each** *a* in *NON-HA* **do**
   *priority(a)* = leaf-node-operator-value(*leaf-node*,*a*)
   **if** *priority(a) > max-HA-priority* **then**
      *(selected(a),rejected(a))* = classify-bindings-tree(*T*, *ℋ*, *a*)
      *selection_ratio(a)*=$\frac{selected(a)}{selected(a)+rejected(a)}$
      *priority(a)* = *priority(a)* + *selection_ratio(a)*
      *selected-actions* = *selected-actions* ∪{*a*}
**return** sort(*selected-actions*, *priority*)

---

Figure 10: Algorithm for ordering actions using relational decision trees.

classification tree in Figure 4). The number of examples covered gives an operator ordering that can be used to prefer actions during the search. The algorithm uses this number to initialize the *priority* value for each helpful action, taking the value of the corresponding operator. The algorithm keeps only *helpful actions* that have at least one matching example. For each of these actions, the algorithm matches the action with its corresponding binding classification tree. The resulting leaf of the binding tree returns two values: the number of times the ground action was *selected*, and the number of times it was *rejected* in the training phase. We define the selection ratio for the ground action as:

$$selection\_ratio(a) = \frac{selected(a)}{selected(a) + rejected(a)}$$

This ratio represents the proportion of good bindings covered by a particular leaf of the binding tree. When the denominator is zero, the selection ratio is assumed to be zero. The priority of the action is updated by adding this selection ratio. Thus, the final priority for an action is higher for actions with operators for which the operator classification tree provides higher values, i.e. they have been selected more often in the training examples. Since the selection ratio remains between 0 and 1, adding up to this number can be considered as a method for breaking ties in the initial priority value, using the information in the binding classification tree.

The priority for non-*helpful actions* is computed in a similar way except that, in this case, the algorithm only considers actions whose initial priority (the value provided by the operator classification tree), is higher than the maximum priority of the *helpful actions*. In this manner, we capture useful non-*helpful actions*. FF follows a heuristic criterion to classify an action as *helpful*. Although this heuristic has been shown to be very useful, the case may arise in which the most useful action at a particular moment is not classified as *helpful*. Decision trees capture this information, given that they can recommend choosing a non-*helpful action*. The described method takes advantage of this fact and defines a way of using such information when applying the learned knowledge. An alternative approach would be to extend the planning context with a new meta-predicate for non-*helpful actions*. However, it does not pay off in a variety of problems and domains because it means significantly larger contexts, which causes more expensive matching. Finally, the selected actions are sorted in order of decreasing priority values. The sorted list of actions is the output of the algorithm.

### 3.4.2 THE H-CONTEXT POLICY ALGORITHM

The *helpful context*-action policy algorithm moves forward, applying at each state the best action according to the DCK. The pseudo-code of the algorithm is shown in Figure 11. The algorithm maintains an ordered *open-list*. The *open-list* contains the states to be expanded which are extracted in order. Once extracted, each state is evaluated using the FF heuristic. Thus, we evaluate upon extraction and not when nodes are included in the *open-list*. The evaluation provides the heuristic value for the state, $h$, and the set of helpful actions $HA$, which are needed to generate the helpful context. The heuristic value is only used for: (1) continuing the search when the state is a recognized dead-end ($h = \infty$), and (2) goal checking ($h = 0$). Then, the helpful context is generated. Subsequently, the algorithm obtains the set $AA$ of actions applicable in the state and sorts them using the decision trees (as shown above in the algorithm in Figure 10). The result is $AA' \subseteq AA$, a sorted list of applicable actions. The algorithm inserts the successors generated by actions in $AA'$ at the beginning of the *open-list* preserving their ordering (function `push-ordered-list-in-open`). Furthermore, to make the algorithm complete and more robust, successors generated by applicable actions that are not in $AA'$ are included in a secondary list called *delayed-list*. The *delayed list* is only used when the *open-list* is empty. In that case, only one node of the *delayed-list* is moved to the *open-list* and then, the algorithm continues extracting nodes from the *open-list*.

In this algorithm, each node maintains a pointer to each parent in order to recover the solution once it has been found. Also, each node maintains its $g$ value, i.e. the length of the path from the initial state up to the node. The function `push-ordered-list-in-open` only inserts in the open list those candidates that: (1) are not repeated states, or (2) are repeated states with lower $g$ value than the previous one. Otherwise, repeated states are pruned. This type of pruning guarantees that we maintain for each node the shortest solution found.

In other words, the proposed search algorithm is a depth-first search with delayed successors. The benefit of this algorithm is that it exploits the best action selection when the policy is per-

---

**Depth-First H-Context Policy** $(I, G, T)$: *plan*

---

$I$: Initial state
$G$: Goals
$T$: Decision Trees

---

*open-list* = $\{I\}$;
*delayed-list* = $\emptyset$;
**while** *open-list* $\neq \emptyset$ **do**
    $n$ = pop(*open-list*)
    $(h, HA)$ = evaluate$(n, G)$  */\*compute FF heuristic\*/*
    **if** $h = \infty$ **then**  */\*recognized dead-end\*/*
        **continue**
    **if** $h = 0$ **then**  */\*goal state\*/*
        **return** path$(I, n)$
    $\mathcal{H}$ = helpful-context$(HA, G, n)$
    *AA* = applicable-actions$(n)$
    *AA'* = DT-Filter-Sort(*AA*, $\mathcal{H}$, $T$)
    *candidates* = generate-successors($n$, *AA'*)
    *open-list* = push-ordered-list-in-open(*candidates*,*open-list*)
    *delayed-candidates* = generate-successors($n$, $AA \setminus AA'$)
    *delayed-list* = push-ordered-list(*delayed-candidates*, *delayed-list*)
    **while** *open-list* $= \emptyset$ **and** *delayed-list* $\neq \emptyset$ **do**
        *open-list* = { pop(*delayed-list*) }
**return fail**

---

Figure 11: A depth-first algorithm with a sorting strategy given by the DCK.

fect[1] and the action ordering when is not. Particularly, perfect DCK will be directly applied in a backtrack-free search and inaccurate DCK will force the search algorithm to backtrack.

### 3.4.3 THE H-CONTEXT POLICY IN A LOOKAHEAD STRATEGY

In many domains the learned DCK may contain flaws: the helpful context may not be expressive enough to capture good decisions, the learning algorithm may not be able to generalize well or the training examples may not be representative enough. In these cases, a direct application of the learned DCK (without backtracking) may not allow the planner to reach the goals of the problem.

Poor quality in the learned DCK can be balanced with a guide of a different nature such as a domain-independent heuristic. A successful example is the *ObtuseWedge* system (Yoon et al., 2008) that combined a learned generalized policy with the FF heuristic. *ObtuseWedge* exploited the learned policy to synthesize *lookahead states* within a lookahead strategy. *Lookahead states* were first applied in heuristic planning by the YAHSP planner (Vidal, 2004). They are intermediate states that are frequently closer to a goal state than the direct descendants of the current state. These intermediate states are added to the list of nodes to be expanded so that they can be used within different search algorithms. When the learned policy contains flaws, *lookahead states* synthesized with the policy may not provide good guidance for the search. However, if these *lookahead states* are included in a complete search algorithm that also considers ordinary successors, the search process becomes more robust. In general, the use of *lookahead states* in a forward state-space search slightly increases the branching factor of the search process, but overall, as shown by the YAHSP planner at IPC-2004 and in the experiments included in the YAHSP paper (Vidal, 2004), the approach seems to improve the performance significantly.

---

1. With perfect policy we refer to a policy that leads directly to a goal state. Our policies are not guaranteed to be perfect given that they are generated by inductive learning.

Figure 12 shows a generic algorithm for using lookahead states generated from a policy during the search. This algorithm is a weighted Best-First Search (BFS), with the only modification being that one or more *lookahead states* are inserted into the open list when expanding a node. As in weighted BFS, nodes to be expanded are maintained in an open list ordered by the evaluation function $f(n) = \omega \times h(n) + g(n)$. Apart from the usual arguments of BFS, the algorithm receives the policy ($P$) and the horizon. The horizon represents the maximum number of policy steps that are applied for generating the lookahead states. In the experiments, we will use this algorithm with the FF heuristic as $h(n)$.

---

**H-Context Policy Lookahead BFS** ($I$,$G$,$T$,$horizon$): *plan*

---

$I$: Initial state
$G$: Goals
$T$: Decision Trees (policy)
$horizon$: horizon

---

*open-list* = $\emptyset$
add-to-open($I$)
**while** *open-list* $\neq \emptyset$ **do**
   $n$ = pop(*open-list*)
   **if** goal-state($n, G$)
      **return** path($I, n$)
   add-to-open-lookahead-successors($n, G, T, horizon$)
   add-to-open-standard-successors($n$)
**return fail**

---

Figure 12: A Generic Lookahead BFS algorithm.

The heuristic evaluation, $h(n)$, the g-value, $g(n)$, and the set of helpful actions, are also saved at each node when the node is evaluated. The function `add-to-open(state)` evaluates the state and inserts it in the *open-list*, which is ordered in increasing values of the evaluation function, $f(n)$. This function also prunes repeated states, following the strategy described for the *Depth-First H-Context Policy* algorithm: only repeated states with higher $g(n)$ than the existent one are pruned. The function `add-to-open-standard-successors(n)` calls `add-to-open` for each successor of the node $n$. The function `add-to-open-lookahead-successors` is explained below.

We have adapted the generic Lookahead BFS algorithm to our learned DCK. Our particular instantiation of the function `add-to-open-lookahead-successors` is shown in Figure 13. In our case, lookahead states are generated by iteratively applying the first action in the action ordering provided by the DCK. The inputs to the algorithm are the current state, the problem goals, the decision trees and the horizon. First, our algorithm generates the helpful context and the applicable actions. The helpful actions, $n.HA$, are recovered from the node. Then, the algorithm sorts the applicable actions using the decision trees (as previously shown in the algorithm in Figure 10). After that, the successor generated by the first action is inserted in the open list, and there is a recursive call with this successor and the horizon decremented by one. The function `add-to-open` returns `true` when its argument has been added to the open list and `false` otherwise. In fact, it returns `false` in only two cases: (1) the state is a repeated state with g-value higher than the g-value of the existent state[2] or (2) the state is a recognized dead-end. When the ordered list becomes empty, the lookahead state can not be generated and the initial node is returned. The same occurs when the horizon is zero. The described implementation is similar to the lookahead strategy approach followed by *ObtuseWedge*, but instead we perform the lookahead generation using helpful contexts and relational decision trees.

On the other hand, in the described *H-Context Policy Lookahead BFS* algorithm the search is perfomed over the set of applicable actions of each node. However, in many domains the use of helpful actions has shown to be a very good heuristic. One possible way of prioritizing helpful actions over non-helpful actions is to include in the open list only those successors given by helpful actions, and to include the remaining successors in a secondary list. We have implemented this idea following the same strategy used in the *Depth-first H-Context Policy* algorithm: when the open list becomes empty only one node is passed from the secondary list to the open list, and the search continues. The algorithm is still complete given that we do not prune any successor. When helpful actions are good enough, this strategy can save many heuristic evaluations. In the experiments we will compare this strategy with the previous one. Our intuition is that the adequacy of each strategy depends directly on the quality of the helpful actions, the quality of the learned DCK, and the accuracy of the heuristic for each particular domain.

Another technique for prioritizing helpful actions in BFS was implemented in YAHSP (Vidal, 2004) which inserts two consecutive instances of each node in the open list. These nodes have equal $f(n)$ since they represent the same state. The first one contains only the helpful actions, and therefore, when expanded, it only generates successors resulting from these actions. The second contains only non-helpful actions, called *rescue actions*. In this way, all the successors with lower $f(n)$ than the parent node in the sub-tree generated by helpful actions are expanded before any successor resulting from non-helpful actions.

We have performed some preliminary experiments, obtaining similar results for the two described methods for prioritizing helpful actions in BFS: the use of a secondary list for non-helpful actions, and the use of rescue nodes. For this reason, we only include results of the first technique in the experimental section. We call this algorithm *H-Context Policy Lookahead BFS-HA*.

---

2. When the state is repeated but with a g-value smaller than the existent one, add-to-open does not re-evaluate but instead takes the heuristic evaluation from the existent state.

---

**add-to-open-lookahead-successors** ($n$,$G$,$T$,$horizon$) :$state$

---

$n$: Node (state)
$G$: Goals
$T$: Decision Trees (policy)
$horizon$: horizon

---

**if** $horizon = 0$ **then**
    **return** $n$
$\mathcal{H}$ = helpful-context($n.HA, G, n$)
$AA$ = applicable-actions($n$)
$AA'$ = DT-Filter-Sort($AA, \mathcal{H}, T$)
**while** $AA' \neq \emptyset$ **do**
    $a$ = pop($AA'$)
    $n'$ = generate-successor($n, a$)
    added = add-to-open($n'$)
    **if** added **then**
        **if** goal-state($n', G$)
            **return** $n'$
        **return** add-to-open-lookahead-successors($n', G, T, horizon - 1$)
**return** $n$

---

Figure 13: Algorithm for generating lookahead states from decision trees.

## 4. Experimental Results

In this section we evaluate the performance of the ROLLER system. The evaluation is carried out over a variety of domains belonging to diverse IPCs: Four domains come from the learning track of IPC-2008 (*Gold-miner*, *Matching Blocksworld*, *Parking* and *Thoughtful*). The rest of the domains (*Blocksworld*, *Depots*, *Satellite*, *Rovers*, *Storage* and *TPP*) were selected from among the domains of the sequential tracks from IPC between 2000 and 2008 because they presented different structure and difficulty, and because they have available random problem generators, so that we can automatically build training sets for learning DCK. For each domain, we complete a **training phase** in which ROLLER learns the corresponding DCK and a **testing phase** in which we evaluate the scalability and quality of the solutions found by ROLLER with the learned DCK. Next, we detail the experimental results obtained at each of these two phases. Moreover, for each of the domains we give particular details about training and test sets, the learned DCK and the observed ROLLER performance.

### 4.1 Training Phase

For each domain, we built a training set of thirty randomly generated problems. The size and structure of these problems is further discussed in the particular details given for each domain. As explained in section 3.2, ROLLER generates its training examples solving the problems from the training set with a BFS-BnB search. We set a time-bound of 60 seconds to solve each training problem, discarding those that are not exhaustively explored in this time-bound. Then, ROLLER generates the training examples from the solutions found and builds the corresponding decision trees with the TILDE system (Blockeel & De Raedt, 1998).

To evaluate the efficiency of ROLLER at the training phase we computed the following metrics: the time needed for solving the training problems, the number of training examples generated in this process, the time spent by TILDE in learning the decision trees and the number of leaves of the operator selection tree. This last number gives a clue about the size of the learned DCK. Table 1 shows the results obtained for each domain.

| Domain | Training Time (s) | Learning Examples | Learning Time (s) | Tree Leaves |
|---|---|---|---|---|
| Blocksworld | 836.0 | 2542 | 13.3 | 18 |
| Depots | 456.2 | 493 | 23.1 | 13 |
| Gold-miner | 1156.9 | 126 | 4.5 | 5 |
| Matching-BW | 865.8 | 430 | 12.4 | 23 |
| Parking | 105.8 | 442 | 7.0 | 12 |
| Rovers | 528.3 | 1011 | 13.6 | 24 |
| Satellite | 19.8 | 1702 | 13.4 | 4 |
| Storage | 136.3 | 677 | 5.1 | 6 |
| Thoughtful | 883.4 | 502 | 352.2 | 19 |
| TPP | 995.9 | 560 | 23.3 | 6 |

Table 1: Experimental results of the training process. Training and learning times are shown, as well as the number of training examples, and complexity of generated trees (number of leaves).

ROLLER achieves shorter *Learning Times*, fourth column in Table 1, than the state-of-the-art systems for learning generalized policies (Martin & Geffner, 2004; Yoon et al., 2008). Particularly, while these systems implement ad-hoc learning algorithms that sometimes require hours in order to obtain good policies, our approach only needs seconds to learn the DCK for a given domain. This fact makes our approach more suitable for architectures that need on-line planning and learning processes. However, these learning times are not constant for different domains, because they depend on the number of training examples (in our work, this number is given by the amount of different solutions for the training problems), on the size of the training examples (in our work this size is given by the number and arity of predicates and actions in the planning domain) and how training examples are structured, i.e., whether examples are easily separated by learning or not.

## 4.2 Testing Phase

In the testing phase ROLLER attempts to solve, for each domain, a set of thirty test problems. These problems are taken from the evaluation set of the corresponding IPC. When this evaluation set contains more problems, these thirty problems are the thirty hardest ones. The *Depots* domain is an exception with twenty-two problems, because the evaluation set for this domain at IPC-2002 only contained those twenty-two problems. Three experiments are made for the testing phase. The first one evaluates ROLLER's performance when DCK is learned with all solutions of the training problems or with the ranked solution approach. The second one evaluates the usefulness of the learned DCK and the third one compares ROLLER with state-of-the-art planners. For each experiment we evaluate two different dimensions of the solutions found by ROLLER: the **scalability** and the **quality**. All testing experiments were done using a 2.4 GHz processor with a time-bound of 900 seconds[3] and 6Gb of memory-bound.

### 4.2.1 SOLUTION RANKING EVALUATION

This experiment evaluates the effect of selecting solutions following the approach described in Section 3.3. The ROLLER configurations for this evaluation are:

- **Top-Ranked Solutions**: The *Depth-First H-Context Policy* algorithm using DCK learned with the sub-set of the top ranked solutions. We use this search algorithm, since its performance depends more on the quality of the learned DCK than that of the other algorithms using DCK.

- **All Solutions**: The *Depth-First H-Context Policy* algorithm using DCK learned with all solutions obtained by the BFS-BnB algorithm.

Table 2 shows the number of problems solved by each configuration, also with the time and plan length average computed over the problems solved by both configurations. The number in brackets in the first column is the number of problems solved in common. The *Top-ranked solutions* configuration solved thirty more problems than *all solutions* configuration, mainly due to the difference of 21 problems in the Matching Blocksworld domain.

| | Top-Ranked Solutions | | | All Solutions | | |
|---|---|---|---|---|---|---|
| Domains | Solved | Time | Length | Solved | Time | Length |
| Blocksworld (30) | **30** | **0,62** | **170,0** | **30** | 2,39 | 550,7 |
| Depots (18) | **21** | **0,94** | **489,1** | 18 | 0,97 | 607,3 |
| Gold-miner (30) | **30** | **0,01** | **65,3** | **30** | **0,01** | **65,3** |
| Matching-BW (0) | **21** | — | — | 0 | — | — |
| Parking (30) | **30** | 4,90 | 148,9 | **30** | **2,20** | **56,2** |
| Rovers (27) | 28 | **1,40** | **166,0** | **29** | 31,20 | 355,8 |
| Satellite (28) | **30** | 11,21 | 123,6 | 28 | 11,47 | **121,6** |
| Storage (10) | **15** | **0,00** | **9,0** | 10 | **0,00** | **9,0** |
| Thoughtful (12) | **12** | **1,25** | 249,7 | **12** | 1,28 | 249,2 |
| TPP (30) | **30** | 0,97 | 147,1 | **30** | **0,90** | **132,8** |
| Total | **247** | — | — | 217 | — | — |

Table 2: Problems solved and time and plan length average for the evaluation on ranking solution heuristic.

The effect of selecting solutions varies across domains. For instance, it is quite important regarding the plan quality for Blocksworld, Depots and Rovers. In the Satellite domain the top-ranked solutions allow ROLLER to solve two more problems while maintaining similar time and plan length average. In the Gold Miner domain, selecting solutions is irrelevant because there are few equally good solutions per problem (i.e., the goal is always the single fact of "having the gold") and fairly most of them are top-ranked ones. The Parking domain does not benefit from selecting solutions. Considering the overall results, we think that selecting solutions is a useful heuristic for improving the DCK quality in many domains. In the remaining evaluations we will refer to DCK used by ROLLER as the decision trees learned with the top-ranked solutions.

---

3. 900 seconds was the time-bound established at the learning track of IPC-2008.

### 4.2.2 DCK USEFULNESS EVALUATION

As shown in IPC Learning Track results, DCK may degrade the performance of the base planner, when the DCK is incorrect. With this in mind, we designed this experiment to measure the performance of ROLLER algorithms comparing them with versions without DCK. We made two versions for the non-learning algorithms. The first one is an *empty* configuration where there is no decision tree given to the algorithm, thus no ordering is computed for the helpful actions, and the second one is the *systematic* configuration, where the ordering is supplied by the FF heuristic instead.

The ROLLER configurations used for the comparisons are:

- **ROLLER**: The *Depth-First H-Context Policy* algorithm with the DCK learned at the training phase.

- **ROLLER-BFS**: The *H-Context Policy Lookahead BFS* algorithm with the DCK learned at the training phase. This configuration uses the horizon $h = 100$. We choose this value on the basis of empirical evaluations.

- **ROLLER-BFS-HA**: A modified version of ROLLER-BFS where only helpful actions are considered as immediate successors. The lookahead states are generated as in the original version, using also the same horizon.

These three algorithms have their equivalent version for the empty configuration:

- **DF-HA** (*Depth-first Helpful Actions*): An empty DCK for ROLLER corresponds to a depth-first algorithm over the helpful actions. As in the original algorithm, non-helpful actions are placed in the delayed list.

- **BFS**: An empty DCK for ROLLER-BFS does not generate lookahead states (i.e., the algorithm *add-to-open-lookahead-successors* in Figure 12). Therefore, the algorithm becomes the standard Best-first Search.

- **BFS-HA**: A modified version of BFS where only helpful actions are considered. Non-helpful actions are placed in a delayed list.

Previous configurations also have a systematic version. In this case action ordering is computed with the FF heuristic:

- **GR-HA** (*Greedy Helpful Actions*): This algorithm corresponds to a greedy search over the helpful actions. For each node, helpful immediate successors are sorted with the FF heuristic. Non-helpful nodes go to the delayed list.

- **LH-BFS** (*Lookahead-BFS*): A BFS with lookahead states. The function *DT-Filter-Sort* is replaced by a function that computes the ordering using the FF heuristic.

- **LH-BFS-HA**: A modified version of LH-BFS where only helpful actions are considered. Non-helpful actions are placed in a delayed list.

For the comparison, we computed the number of problems solved and the scores used in the IPC-2008 learning track to evaluate planners performance in terms of CPU time and quality (plan length). The time score is computed as follows: for each problem $i$ the planner receives $T_i^*/T_i$ points, where $T_i^*$ is the minimum time a participant used for solving the problem $i$, and $T_i$ is the CPU time used by the planner in question. In a 30 problem test set a planner can receive at most 30 points, the higher the score the better. The quality score is computed in the same way, just replacing $T$ with $L$, where $L$ measures the quality in terms of plan length. In addition we compute the time and quality averages for problems solved by all configurations. If a configuration did not solve any problem, it is not taken into account for this measure. Average measures complement scores since they give a direct information for commonly solved problems, while scores tend to benefit configurations that solve problems which others do not.

Table 3 shows a summary for the results obtained in the DCK usefulness evaluation. For each configuration we compute the number of domains where the algorithm was the top performer for each of the evaluated criteria (i.e., numbers of solved problems, time and quality scores and averages). A top performer in a domain is an algorithm that has equal to or better measure than the rest of the algorithms. In the table, each algorithm can have at most 10 points, the number of evaluated domains. *Global* section refers to overall top performers. *Relative* section refers to the number of domains where a configuration was equal or better than the other two configurations of the same algorithm strategy (i.e., depth-first, best-first, best-first with helpful actions). All averages of commonly solved problems were computed for configurations that solve more than one problem. Results show that ROLLER is very good in the number of solved problems and speed metrics. Regarding quality score, ROLLER and ROLLER-BFS-HA were the best performers in three domains each. However, BFS and BFS-HA obtained better results in quality average.

| | DEPTH-FIRST | | | BEST-FIRST | | | HELPFUL BEST-FIRST | | |
|---|---|---|---|---|---|---|---|---|---|
| *Global* | roller | gr-ha | df-ha | roller-bfs | lh-bfs | bfs | roller-bfs-ha | lh-bfs-ha | bfs-ha |
| Solved Problems | 7 | 2 | 2 | 1 | 0 | 1 | 5 | 1 | 1 |
| Time Score | 8 | 1 | 0 | 0 | 0 | 0 | 1 | 0 | 0 |
| Time Average | 9 | 1 | 1 | 1 | 0 | 0 | 1 | 0 | 0 |
| Quality Score | 3 | 1 | 0 | 1 | 0 | 1 | 3 | 1 | 2 |
| Quality Average | 0 | 0 | 0 | 1 | 2 | 3 | 0 | 1 | 5 |
| *Relative* | | | | | | | | | |
| Solved Problems | 8 | 3 | 2 | 7 | 3 | 2 | 9 | 3 | 2 |
| Time Score | 9 | 1 | 0 | 8 | 1 | 1 | 9 | 0 | 1 |
| Time Average | 9 | 1 | 1 | 9 | 1 | 0 | 7 | 1 | 2 |
| Quality Score | 7 | 3 | 0 | 4 | 4 | 2 | 7 | 1 | 2 |
| Quality Average | 5 | 5 | 0 | 1 | 2 | 7 | 2 | 1 | 7 |

Table 3: Summary for DCK usefulness evaluation. Each column gives the number of domains where each configuration was the top performer for a row item.

Table 4 shows the number of solved problems for the DCK usefulness evaluation. The *Total* row shows that each ROLLER configuration solved more problems than the empty and systematic versions. Results for time and quality scores are reported in Table 9 and Table 10 of Appendix A.

Detailed results for averages were considered less interesting since in many domains there are very few common solved problems, which are the easy problems.

| | DEPTH-FIRST | | | BEST-FIRST | | | HELPFUL BEST-FIRST | | |
|---|---|---|---|---|---|---|---|---|---|
| Domains | roller | gr-ha | df-ha | roller-bfs | lh-bfs | bfs | roller-bfs-ha | lh-bfs-ha | bfs-ha |
| Blocksworld (30) | **30** | 1 | 0 | 8 | 0 | 0 | 8 | 0 | 0 |
| Depots (22) | **21** | 18 | 18 | 20 | 19 | 13 | 20 | 20 | 20 |
| Gold-miner (30) | **30** | 0 | 0 | 17 | 17 | 16 | **30** | 0 | 0 |
| Matching-BW (30) | **21** | 0 | 0 | 14 | 7 | 14 | 19 | 10 | 17 |
| Parking (30) | **30** | 25 | 1 | **30** | 11 | 7 | **30** | 11 | 9 |
| Rovers (30) | 28 | **30** | **30** | 26 | 28 | 11 | **30** | **30** | **30** |
| Satellite (30) | **30** | 23 | 22 | 25 | 22 | 15 | **30** | 23 | 23 |
| Storage (30) | 15 | 9 | 10 | 19 | 18 | **20** | 19 | 10 | 10 |
| Thoughtful(30) | 12 | 15 | 0 | 20 | 14 | 11 | **23** | 16 | 12 |
| TPP (30) | **30** | 30 | 30 | 16 | 24 | 9 | 19 | 26 | 14 |
| Total | **247** | 151 | 111 | 195 | 160 | 116 | 228 | 146 | 135 |

Table 4: Problems solved for the DCK usefulness evaluation.

### 4.2.3 TIME PERFORMANCE COMPARISON

This experiment evaluates the scalability of the ROLLER system, compared to state-of-the-art planners. For the comparison, we have chosen LAMA (Richter & Westphal, 2010), the winner of the sequential track of the past IPC, and FF, which in the last IPC has shown to be still competitive. We used the three ROLLER configurations explained in the previous evaluation. The configuration for other planners are:

- **FF**. Running the Enforced Hill-Climbing (EHC) algorithm with helpful actions together with a complete BFS in case EHC fails [4]. Though this planner dates from 2001 we include it in the evaluation because, as shown by the results of IPC-2008, it is still competitive with other state-of-the-art planners. Besides, this planner is extensively used in other planning and learning systems.

- **LAMA-first**. The winner of the classical track of IPC-2008. In this configuration LAMA is modified to stop when it finds the first solution. In this way, comparison is fair because the rest of configurations do not implement *anytime* behavior, i.e., the continuous solution refinement until reaching the time-bound). The *anytime* behavior of LAMA is compared later with the ROLLER performance in the next section.

Table 5 shows the number of problems solved together with the speed score. These results give an overall view of the performance of the different planners. ROLLER solves as many or more problems than any other configuration in 6 of 10 domains and achieves the top speed score in seven domains. The second best score belongs to ROLLER-BFS-HA, which solves as many or more problems than other planners in six domains. LAMA-first is fairly competitive, since it solves seven problems less than ROLLER and 13 more problems than ROLLER-BFS-HA. In both cases LAMA-first achieves a lower speed score.

---

4. This planner is actually Metric-FF running STRIPS domains. We consider this implementation an adequate baseline for comparison because ROLLER was implemented over this code rather than over the original FF in order to extend our approach to other planning models.

| Domain (problems) | ROLLER | | ROLLER-BFS | | ROLLER-BFS-HA | | FF | | LAMA-first | |
|---|---|---|---|---|---|---|---|---|---|---|
| | solved | score | solved | score | solved | score | solved | score | solved | score |
| Blocksworld (30) | **30** | **29.87** | 8 | 2.47 | 8 | 2.40 | 0 | 0.00 | 17 | 0.17 |
| Depots (22) | **21** | **19.86** | 20 | 11.01 | 20 | 11.46 | 20 | 8.70 | 20 | 3.88 |
| Gold-miner (30) | **30** | **26.00** | 17 | 0.03 | **30** | 5.35 | 27 | 0.22 | 29 | 12.24 |
| Matching-BW (30) | 21 | 14.84 | 14 | 1.32 | 19 | 1.71 | 9 | 0.27 | **25** | **20.02** |
| Parking (30) | **30** | **28.57** | **30** | 22.72 | **30** | 23.60 | 24 | 0.94 | 23 | 1.69 |
| Rovers (30) | 28 | 24.82 | 26 | 13.41 | **30** | 16.24 | 29 | 5.51 | **30** | 18.59 |
| Satellite (30) | **30** | **22.60** | 25 | 14.61 | **30** | 18.33 | 22 | 5.31 | 28 | 15.66 |
| Storage (30) | 15 | 11.02 | **19** | 12.31 | **19** | **16.17** | 17 | 10.51 | **19** | 9.03 |
| Thoughtful(30) | 12 | 11.99 | 20 | 12.38 | **23** | **13.09** | 14 | 8.16 | 20 | 11.29 |
| TPP (30) | **30** | **29.50** | 16 | 14.83 | 19 | 13.97 | 26 | 6.27 | **30** | 9.66 |
| Total | **247** | **219.07** | 195 | 105.09 | 228 | 122.32 | 188 | 45.89 | 241 | 102.23 |

Table 5: Problems solved and speed score of the five configurations.

Table 6 shows the average time for the five configurations when addressing the subset of problems solved by all configurations. The first column shows in parenthesis the number of commonly solved problems. These results are closely related to those shown in Table 5. ROLLER achieves the best average time in eight out of ten domains. We also observe that different configurations are good in particular domains and even more so in particular problems. For instance, in the *Thoughtful* domain there were only four problems solved by all the configurations.

| Domain (problems) | ROLLER | ROLLER-BFS | ROLLER-BFS-r | FF | LAMA-first |
|---|---|---|---|---|---|
| Blocksworld (7) | **0.36** | 66.31 | 67.99 | | 139.69 |
| Depots (18) | **0.84** | 15.53 | 2.54 | 4.01 | 61.73 |
| Gold-miner (17) | **0.00** | 49.82 | 0.02 | 0.28 | 0.01 |
| Matching-BW (6) | 1.99 | 42.25 | 44.53 | 74.90 | **1.96** |
| Parking (22) | **1.86** | 2.91 | 2.78 | 74.02 | 108.22 |
| Rovers (25) | **1.37** | 24.38 | 9.83 | 42.82 | 1.59 |
| Satellite (22) | **1.24** | 7.08 | 1.87 | 18.23 | 1.33 |
| Storage (14) | 11.74 | **0.01** | 0.03 | 0.05 | 0.19 |
| Thoughtful(4) | **1.49** | 10.84 | 9.52 | 14.52 | 3.55 |
| TPP (16) | **0.02** | **0.02** | **0.02** | 0.70 | 0.10 |

Table 6: Planning time averages in the problems solved by all the configurations.

### 4.2.4 Quality Performance Comparison

This experiment compares the quality of the first solutions found and the solutions found by the *anytime* behavior. In the anytime configuration, planners exhaust the time-bound trying to improve incrementally the best solution found. Three ROLLER algorithms are modified to a configuration where the best solution found so far is used as an upper-bound in order to prune all nodes that exceed this plan length. The anytime behavior is the regular configuration for LAMA. FF does not have anytime behavior, but it will be included in the anytime comparison as well as a base for comparing quality improvements of other planners.

Table 7 shows the quality scores for the first solution and for the last solution found by the anytime configurations. The anytime column for each planner shows the score variation and reveals whether or not the planner was able to make relative improvements of the first solutions. The *relative* for FF shows the score of its solutions compared to the solutions given by the anytime configuration of other planners. FF loses points in most cases because the others were able to improve their solutions. The two LAMA configurations obtained the top score in their category. Nevertheless, no planner dominated in all the domains. Furthermore, all configurations achieved the top quality score for the first solution in at least one domain.

| Domain | ROLLER | | ROLLER-BFS | | ROLLER-BFS-HA | | FF | | LAMA | |
|---|---|---|---|---|---|---|---|---|---|---|
| | first | anytime | first | anytime | first | anytime | first | relative | first | anytime |
| Blocksworld | **29.83** | **29.83** | 8.00 | 8.00 | 8.00 | 8.00 | 0.00 | 0.00 | 7.42 | 8.29 |
| Depots | 8.50 | 9.26 | 12.39 | 17.01 | 12.85 | 18.95 | **19.01** | 17.96 | 18.32 | **19.28** |
| Gold-miner | 14.30 | 18.00 | 11.50 | 17.00 | 13.08 | 15.39 | **27.00** | **27.00** | 14.04 | 26.81 |
| Matching-BW | 9.43 | 9.52 | 13.01 | 12.43 | 17.67 | 17.16 | 8.23 | 7.15 | **23.25** | **24.72** |
| Parking | 19.38 | 17.04 | **24.24** | 23.98 | **24.24** | **25.51** | 21.53 | 17.79 | 19.16 | 22.56 |
| Rovers | 21.38 | 21.39 | 21.78 | 21.59 | 25.66 | 26.14 | **28.66** | 28.33 | 28.26 | **28.97** |
| Satellite | **28.65** | 28.81 | 23.20 | 23.00 | 28.18 | **28.94** | 21.55 | 21.33 | 27.02 | 27.42 |
| Storage | 13.41 | 13.46 | 15.69 | 18.38 | 15.64 | 17.26 | 16.23 | 15.80 | **17.24** | **18.81** |
| Thoughtful | 6.27 | 6.21 | 15.93 | 15.12 | 18.63 | 18.35 | 13.96 | 13.09 | **18.84** | **18.59** |
| TPP | 25.38 | 24.26 | 14.45 | 15.09 | 16.80 | 17.77 | 23.42 | 21.56 | **29.99** | **29.82** |
| Total | 176.53 | 177.35 | 160.19 | 171.65 | 180.75 | 193.53 | 179.59 | 170.06 | **203.54** | **219.24** |

Table 7: Quality scores for the first solution and the anytime configuration of evaluated planners.

| Domain | ROLLER | | ROLLER-BFS | | ROLLER-BFS-HA | | FF | | LAMA | |
|---|---|---|---|---|---|---|---|---|---|---|
| | first | anytime | first | anytime | first | anytime | first | relative | first | anytime |
| Blocksworld (7) | 146.29 | 146.29 | **142.86** | **142.86** | **142.86** | **142.86** | – | – | 358.57 | 318.00 |
| Depots (18) | 385.78 | 372.22 | 81.78 | 54.00 | 76.83 | 43.33 | **46.39** | 46.39 | 49.28 | **41.56** |
| Gold-miner (17) | 55.65 | 38.18 | 30.06 | **19.65** | 47.88 | 39.35 | **19.65** | **19.65** | 43.35 | **19.65** |
| Matching-BW (6) | 186.00 | 170.33 | 75.00 | 70.00 | 76.00 | 69.33 | **71.67** | 71.67 | 78.33 | **62.33** |
| Parking (22) | 96.91 | 93.82 | 75.32 | 59.86 | 75.32 | 54.45 | **60.00** | 60.00 | 64.14 | **47.91** |
| Rovers (25) | 150.80 | 149.96 | 115.56 | 115.56 | 114.40 | 112.12 | **94.20** | 94.20 | 101.44 | 98.36 |
| Satellite (22) | 78.41 | 77.59 | 80.05 | 80.00 | 80.05 | 77.32 | 77.18 | 77.18 | **76.91** | **75.50** |
| Storage (14) | 43.07 | 42.64 | 15.21 | 11.36 | 15.64 | 13.29 | **12.43** | 12.43 | 12.71 | **11.29** |
| Thoughtful(4) | 292.25 | 291.75 | 168.50 | 168.25 | 168.50 | 164.50 | **123.25** | **123.25** | 140.25 | 128.50 |
| TPP (16) | 60.25 | 57.00 | 60.00 | 52.06 | 60.25 | 49.38 | 59.19 | 59.19 | **51.81** | **47.94** |

Table 8: Quality averages for the first solution and the anytime configuration of evaluated planners.

Table 8 shows the plan length average for the problems solved by all configurations. The *first* column shows the average for the first solutions and the *anytime* column gives the average for the last solutions of the anytime configuration. The commonly solved problems are the same as those reported in Table 6. Although FF is the planner that solved fewer problems, it achieves the best average plan length in seven domains. Plan length averages reveal that ROLLER is not able to find first solutions of good quality for most domains. ROLLER-BFS and ROLLER-BFS-HA find better quality solutions than ROLLER, and in several domains, their averages are competitive with LAMA. ROLLER-BFS and ROLLER-BFS-HA show a better quality performance mainly due to the combination of learned DCK with a domain-independent heuristic within the BFS algorithm.

In the following subsections we discuss particular details for each of the domains. We give a very brief description of the domain together with information about training and test sets used in the experimental evaluation. For each domain, we also analyze the learned DCK and the obtained results in order to give a fine-grained interpretation of the observed performance. Further details on these domains can be found at the IPC web site.[5]

### 4.2.5 BLOCKSWORLD DETAILS

Problems in this domain are concerned with configuring towers of blocks using a robotic arm. The training set used for the experiments consisted of: ten eight-block problems, ten nine-block problems and ten ten-block problems. The test set consisted of the 30 largest typed problems from IPC-2000, which have from 36 to 50 blocks.

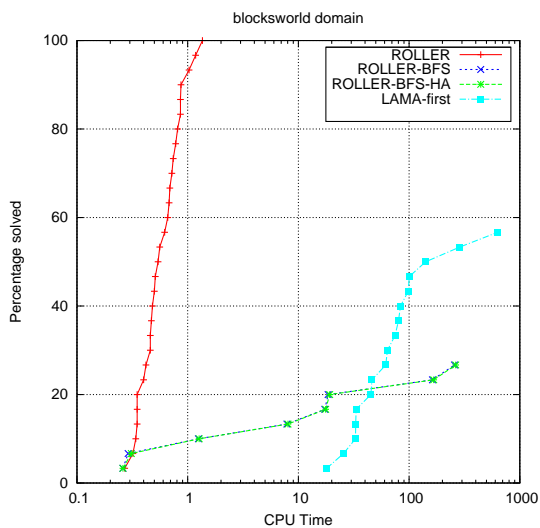

Figure 14: Percentage of solved problems when increasing time for evaluating the scalability performance in the *Blocksworld* domain.

Although this domain is one of the oldest benchmarks in automated planning, it is still challenging for state-of-the-art heuristic planners. *Blocksworld* presents strong interaction among goals that current heuristics fail to capture. In particular, achieving a goal in this domain may undo previously satisfied goals. Therefore, it is crucial to achieve goals in a specific order. The DCK learned by ROLLER gives a total order of the domain actions in different contexts capturing this key knowledge, which lets ROLLER achieve impressive scalability results while producing good quality solution plans. ROLLER configurations are considerably better than non-learning configurations. Particularly, ROLLER solved the thirty problems of the set while DF-HA and GR-HA did not solve any problem. ROLLER is also quite good when compared to state-of-the-art planners. In Figure 14 we can observe that ROLLER performs two orders of magnitude faster than LAMA. The *x-axis* of the figure represents the CPU time in logarithmic scale and the *y-axis* represents the percentage of solved problems in a particular time. Moreover, ROLLER obtained the best quality score in the *first solution* and *anytime* evaluations. In addition, the average plan length of common problems is fairly close to the best average, obtained by ROLLER-BFS and ROLLER-BFS-HA. BFS algorithms do not scale well in this domain because they are partially guided by the FF heuristic, which considerably underestimates the distance to the goals. Similarly, lookahead states generated by the policy are discarded because they fail to escape *plateaus* generated by this heuristic function.

When analyzing the learned operator tree we found explanations for the good performance of ROLLER in the *Blocksworld* domain: The operator tree is clearly split in two parts. The first part contains decisions to take when the arm is holding a block. In this situation, the tree captures when to STACK or PUT-DOWN a block. The second part contains decisions to take when the arm is empty. In this case the tree captures when to UNSTACK or PICK-UP a block. In this second part of the tree, if the current state of the search matches the logical query `helpful_unstack(Block1,Block2)` [6] it means that the tower of blocks under `Block1` is not well arranged, i.e., `Block1` or at least one block beneath `Block1` is not well placed. Therefore, the set of *helpful actions* compactly encodes the useful concept of a *bad tower*. This kind of knowledge was manually defined in previous works in order to learn good policies for *Blocksworld*. One approach consisted of including recursive definitions of new predicates, such as the *support predicates* `above(X,Y)` and `inplace(X)` (Khardon, 1999). Another alternative involved changing the representation language, for instance the *concept language* (Martin & Geffner, 2004) or the *taxonomic syntax* (Yoon, Fern, & Givan, 2007). The Kleene-star operator of the *taxonomic syntax* (i.e., the operator for defining recursion) was discarded in a subsequent work (Yoon et al., 2008) and the `above` predicate was used instead. ROLLER's ability to recognize *bad-towers* without extra predicates arises because any misplaced block in a tower makes the UNSTACK action of the top block *helpful*, since it is always part of the relaxed plan when the arm is empty.

Due to the extraordinary performance of ROLLER in this domain, we built an extra test set to clarify whether or not the trend observed in the ROLLER configuration would hold for larger problems. With this aim, we randomly generated 30 problems distributed in sub-sets of 50, 60, 70, 80, 90 and 100 blocks with 5 problems for each sub-set. ROLLER solved the 30 problems in this extra test set with a time average of 20.1 seconds per problem and spending at most 175.3 to solve a problem. Obviously, problems became more difficult for ROLLER as the number of blocks increase.

### 4.2.6 DEPOTS DETAILS

This domain is a combination of a transportation domain and the *Blocksworld* domain, where there are crates instead of blocks and hoists instead of the robot arm. The problems consist of trucks transporting crates around depots and distributors. Using hoists, crates can be stacked onto pallets or on top of other crates at their final destination. In this domain, the 30 training problems are different combinations of 2 or 3 locations (depots and distributors), 1 or 2 trucks, 1 or 2 pallets per location, 1 hoist per location and from 2 to 5 crates to be placed in different configurations. For the testing phase we have used the 22 problems of the IPC-2002 set. The hardest problem has 12 locations (1 or 2 pallets and 1 or 2 hoists), 6 trucks and 20 crates.

ROLLER and ROLLER-BFS improve the performance of the non-learning strategies, but the three configurations of BFS Helpful-Action solved the same 20 problems. ROLLER is able to solve 21 problems, achieving the best speed score. However, the high average plan length indicates that the policy is not producing good quality plans. ROLLER-BFS-HA obtains the second best speed score with more competitive plan lengths. Figure 15 shows the percentage of solved problems while increasing the CPU time (in logarithmic scale). In the anytime configuration, ROLLER-BFS-HA is able to refine its solutions, achieving a quality average similar to LAMA.

---

6. As explained in section 3.2 logic queries in ROLLER present the example and problem Ids. In this case these Ids are ignored for simplicity given that they are not needed for matching the current helpful context.

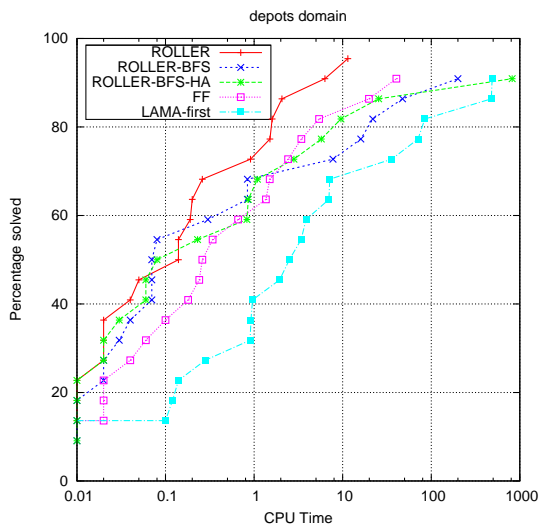

Figure 15: Percentage of solved problem when increasing time for evaluating the scalability performance in the *Depots* domain.

The DCK learned in this domain provides inaccurate advice for large planning contexts. For instance, ROLLER makes some mistakes when deciding which crate to unload when several crates are loaded in a truck. The reason for the inaccurate DCK is that training problems are not large enough to gain this knowledge. In addition, adding more crates to these problems makes it unfeasible for them to be solved with BFS-BnB. Nevertheless, this limitation of the learned DCK is not very evident. The *Depots* domain is *undirected* (i.e., all actions are reversible), so it has no dead ends. Therefore, mistakes made by the DCK are fixed with additional actions, which leads to worse quality plans. Besides, since first solutions are rapidly found, ROLLER configurations can spend time refining solutions. This is the reason for the great improvement in the plan average for ROLLER-BFS-HA.

### 4.2.7 Gold-Miner Details

The objective of this domain is to navigate in a grid of cells until reaching a cell containing gold. Some of the cells are occupied by rocks that can be cleared using bombs or a laser. In this domain the training set consists of: 10 problems with $3 \times 3$ cells, 10 problems with $4 \times 4$ cells, and 10 problems with $5 \times 5$ cells. This domain was part of the learning track in IPC-2008 so we have used the same test set used in the competition. This set has problems ranging from $5 \times 5$ up to $7 \times 7$ cells.

Problems in the *Gold-Miner* domain are not solvable with helpful actions alone. This explains the difference in the number of solved problems between ROLLER, ROLLER-BFS-HA and their non-learning counterpart. In general terms, this domain is trivial for ROLLER, ROLLER-BFS-HA (they solved all the test problems in less than 10 seconds per problem) and LAMA. Nevertheless, FF scales-up poorly. In this domain essential actions for picking up bombs are frequently not considered *helpful actions*, because the relaxed problem is solvable using the laser. Consequently, FF fails to solve most problems with EHC and it requires an additional BFS search. Figure 16 shows the percentage of solved problems while increasing the CPU time. Regarding the anytime evaluation, all tested configurations improved the first solution found in many problems.

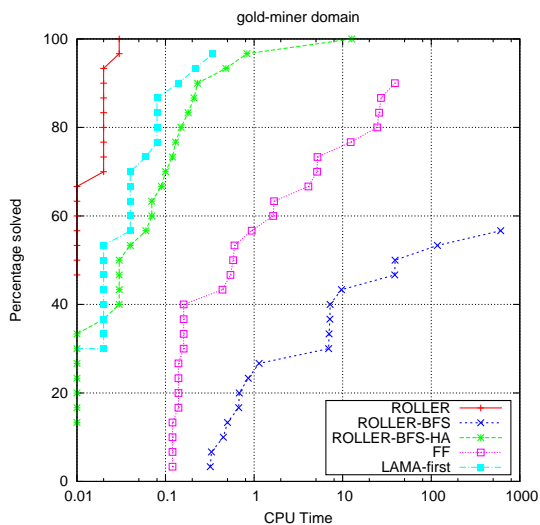

Figure 16: Percentage of solved problems when increasing time for evaluating the scalability performance in the *Gold-miner* domain.

In this domain, the operator tree succeeds in capturing the key knowledge. In the initial states, the bombs and the laser are in the same cell, so the robot needs to decide which of them to pick-up. The operator tree for this domain matches the logical query `candidate_pickup_laser(Cell)` with a higher ratio for operator `PICKUP-BOMB` than for operator `PICKUP-LASER`. This operator preference allows ROLLER to avoid dead ends when the laser destroys the gold. On the other hand, situations where the laser is required (i.e., to destroy hard rocks) are reached as a second choice of the policy. This fact implies some backtracking for ROLLER, but the additional evaluated nodes do not significantly affect the overall performance. The preference of the `PICKUP-BOMB` over the `PICKUP-LASER` action is an example of selecting non-*helpful actions*.

### 4.2.8 MATCHING BLOCKSWORLD DETAILS

This domain is a version of *Blocksworld* designed to analyze limitations of the relaxed plan heuristic. In this version blocks are polarized, either positive or negative. There are also two polarized robot arms. Furthermore, when a block is placed (stack or put-down actions) with an arm of different polarity, the block becomes damaged and no block can be placed on top of it. However, picking up or unstacking a block with the wrong polarity seems to be harmless. This fact makes recognizing dead ends a difficult task for the FF heuristic. Particularly, in the relaxed task blocks are never damaged. Thus, both the relaxed plan (and consequently the set of helpful actions) and the heuristic estimation are wrong. The training set used in this domain consists of fifteen 6-blocks problems and fifteen 8-blocks problems. We used an even number of blocks to keep the problems balanced (i.e., half of the blocks of each polarity). For the testing phase we used the test set from the learning track of IPC-2008. This set has problems ranging from 15 to 25 blocks.

DF-HA and GR-HA did not solve any problem, because these problems are not solvable with helpful actions alone. The learned DCK recommended some useful non-helpful actions, thus ROLLER was able to solve 21 problems. Policy configurations perform better than systematic strategies, but are fairly similar to not using a lookahead strategy. This fact reveals that the learned DCK is not effective enough to pay off the effort of building lookahead states. LAMA is the planner that solves the most problems. Figure 17 shows the percentage of solved problems while increasing the CPU time.

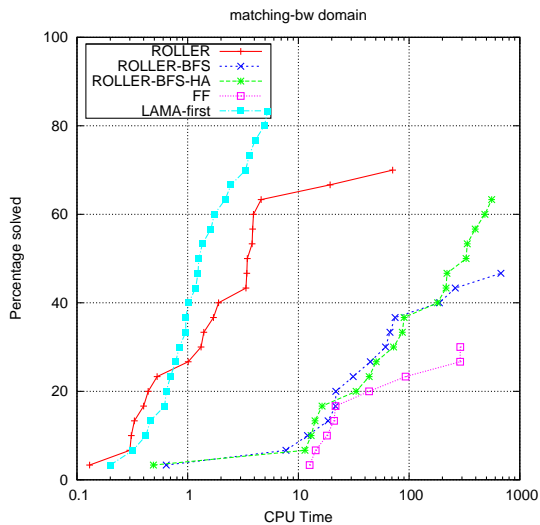

Figure 17: Percentage of solved problems when increasing time for evaluating the scalability performance in the *Matching Blocksworld* domain.

ROLLER solved problems by evaluating a considerable number of nodes above the plan length, which means that the DCK learned for this domain is not accurate. When analyzing the training examples we find many solution plans that do not satisfy the key knowledge of the domain (*robot arms should unstack or pick-up blocks of the same polarity*). Specifically, when the robot is handling a *top block*, i.e., a block with no other blocks above it in the goal state, then the polarity of the robot arm becomes meaningless. This effect is unavoidable because the shortest plans involve managing top blocks in a more efficient way while ignoring the polarities. These examples include noise in the learning and make generalization very complex.

### 4.2.9 PARKING DETAILS

This domain involves parking cars on a street with $N$ curb locations where cars can be double parked, but not triple parked. The goal is to move from one configuration of parked cars to another by driving cars from one curb location to another. In this domain the training set consists of: fifteen problems with six cars and four curbs and fifteen problems with eight cars and five curbs. For testing we used the test set from the learning track of IPC-2008. The hardest problem in this set has 38 cars and 20 curbs.

The three ROLLER configurations solve all problems and perform significantly better than non-learning strategies. In addition, the three ROLLER configurations outperform FF and LAMA with a difference of more than one order of magnitude. This is the reason for LAMA and FF low speed scores. ROLLER configurations are also consistently better than systematic and empty configurations. Figure 18 shows the percentage of solved problems while increasing the CPU time. On the other hand, the three ROLLER configurations did not achieve first solutions of suficient quality. However, these solutions are refined in the anytime evaluation, especially by ROLLER-BFS-HA, which achieves the top quality score and has a plan length average fairly similar to LAMA.

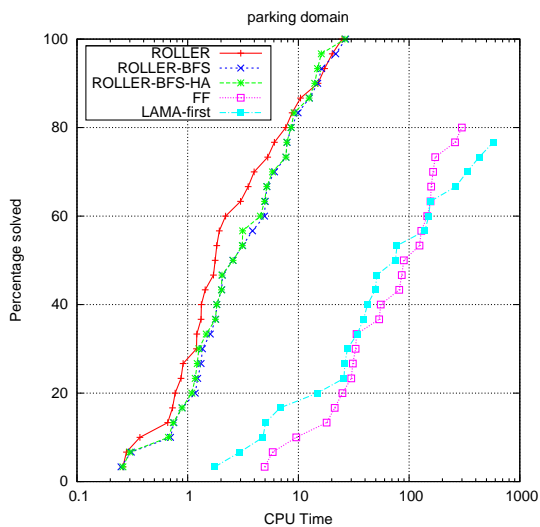

Figure 18: Percentage of solved problems when increasing time for evaluating the scalability performance in the *Parking* domain.

The learned DCK in this domain was quite effective (ROLLER rarely backtracked). The operator tree perfectly classifies the MOVE-CAR-TO-CURB action at the first tree node, asking if it is considered a *helpful action*. Besides, the binding tree for this operator selects the right car by asking about the target goal and rejecting other candidates. These two decisions guide the planner to place a car in the right position whenever possible. As a result, a large number of nodes are not evaluated, which explains the scalability difference with FF and LAMA.

### 4.2.10 ROVERS DETAILS

This domain is a simplification of the tasks performed by the autonomous exploration vehicles sent to Mars. The tasks consist of navigating the rovers, collecting soils and rocks samples, and taking images of different objectives. In this domain the training set consists of: ten problems with one rover, four waypoints, two objectives and one camera; ten problems with an additional camera; and ten problems with an additional rover. Problems in the test set are the thirty largest problems from the IPC-2006 set (i.e., problems 11 to 40). The largest problem in this set has 14 rovers and 100 waypoints.

DCK strategies are faster than systematic and empty strategies, but differences are not significant since all configurations solved most of the problems. On the one hand helpful actions in the *Rovers* domain are quite good but on the other hand the test set does not have problems which are big enough to generate differences among approaches. Regarding planner comparison, ROLLER achieves the top performance score and scales significantly better than FF, and solves two problems less than LAMA. Figure 19 shows the percentage of solved problems when increasing the CPU time. Regarding the anytime evaluation, all planners are able to refine their first solutions. LAMA gets the top quality score and the best plan length after refining solutions.

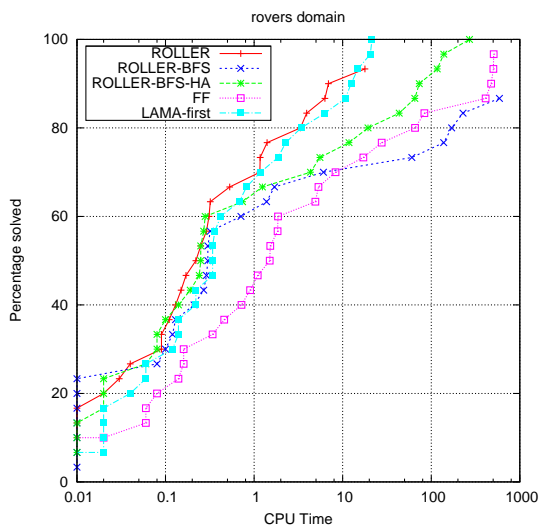

Figure 19: Percentage of solved problems when increasing time for evaluating the scalability performance in the *Rovers* domain.

In this domain, ROLLER learned imperfect DCK, but it manages to achieve good scalability results. DCK is imperfect partially because actions for communicating rock, soil or image analysis can be applied in any order among them. Therefore, the preferences for ranking and selecting solutions fail to discriminate among these actions which confuse the learning algorithm. Since these actions could be applied in any order, the DCK mistakes seem to be harmless at planning time.

### 4.2.11 SATELLITE DETAILS

This domain comprises a set of satellites with different instruments, which can operate in different formats (modes). The tasks consist of managing the instruments for taking images of certain targets in particular modes. In this domain the training set consist of thirty problems with one satellite, two instruments, five modes and five observations. Problems in the test set are the thirty largest problems from the IPC-2004 (i.e., problems 7 to 36). The largest problem in this set has 10 satellites, 5 modes and 174 observations.

The three ROLLER configurations improved the number of solved problems of their non-learning counterpart. In addition, ROLLER and ROLLER-BFS-HA solved the 30 problems in the set, two more than LAMA and eight more than FF. Figure 20 shows the percentage of solved problems when increasing the CPU time. ROLLER and ROLLER-BFS-HA achieve good quality solutions and are able to refine them in the anytime evaluation, achieving plan lengths similar to LAMA.

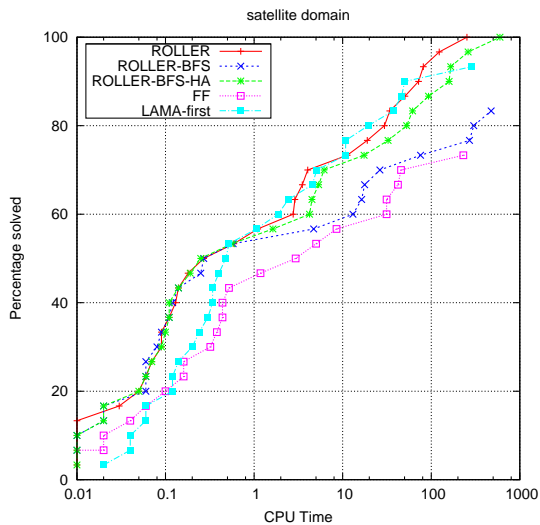

Figure 20: Percentage of solved problems when increasing time for evaluating the scalability performance in the *Satellite* domain.

The learned DCK captures the key knowledge of the *Satellite* domain. The trees shown in Figure 4 and Figure 7 are part of the learned DCK with fewer training examples. In this domain both ROLLER and ROLLER-BFS-HA perform quite similarly. The reason is that the FF heuristic is also quite accurate in the domain. Thus, the deepest lookahead state generated by the learned policy is frequently selected by the heuristic in the BFS search.

### 4.2.12 STORAGE DETAILS

This domain is concerned with the storage of a set of crates taking into account the spatial configuration of a depot. The domain tasks comprise using hoists to move crates from containers to a particular area in the depot. The training set consists of 30 problems with 1 depot, 1 container, 1 hoist and different combinations of 2 or 3 crates and from 2 up to 6 areas inside the depot. For the test set we used the 30 problems from the IPC-2006 set. The largest problem in this domain has 4 depots with 8 areas each, 5 hoist and 20 crates.

The first 12 problems are trivially solved by all configurations. Then, problem difficulty increases quickly when the number of problem objects increases. The BFS solved 20 problems, one more than any DCK strategy, meaning that DCK lookahead strategies do not pay off. This domain is also hard for FF and LAMA. Figure 21 shows the percentage of solved problems when increasing the CPU time.

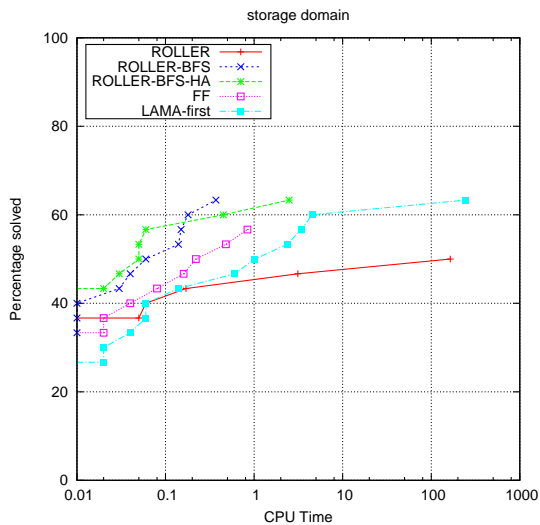

Figure 21: Percentage of solved problems when increasing time for evaluating the scalability performance in the *Storage* domain.

Although DCK is not effective, we found interesting properties in it. The learned operator tree is compact and succeeds in selecting the GO-IN action which is normally not marked as a *helpful action*.

### 4.2.13 THOUGHTFUL DETAILS

This domain models a version of the *solitaire* card game, where all cards are visible and one can turn each card from the talon rather than 3 cards at a time. As in the original version, the goal of the game is to place all cards in ascending order in their corresponding suit stacks (home deck). There is no available random problem generator for this domain. Therefore, we used the bootstrap problem distribution given in the learning track of IPC-2008. This set contains problems for the four suits, having up to card seven for each suit. For the test phase we used the 30 problems from the test distribution of the learning of IPC-2008. The largest problem in this domain has the full set of a standard card game.

ROLLER only solves 12 problems, three fewer than GR-HA. However, ROLLER-BFS and ROLLER-BFS-HA are better in the number of solved problems than non-learning approaches. In this domain, the use of DCK for lookahead construction combined with the FF heuristic makes the search process more robust against policy mistakes. ROLLER-BFS-HA solves 23, three more than LAMA. Figure 22 shows the percentage of solved problems when increasing the CPU time.

The BFS-BnB algorithm for generating training examples is only able to solve 12 out of 30 problems from the bootstrap problem distribution. We believe that a different bootstrap distribution with smaller problems would generate more accurate DCK. Additionally, even though DCK lookahead strategies achieve good results, learning accurate decision trees is more complex when there are many more classes (20 operators in this particular domain) and many more arguments in the predicates of the background knowledge (up to 6 parameters in operator `col-to-home` and 7 parameters in operator `col-to-home-b`).

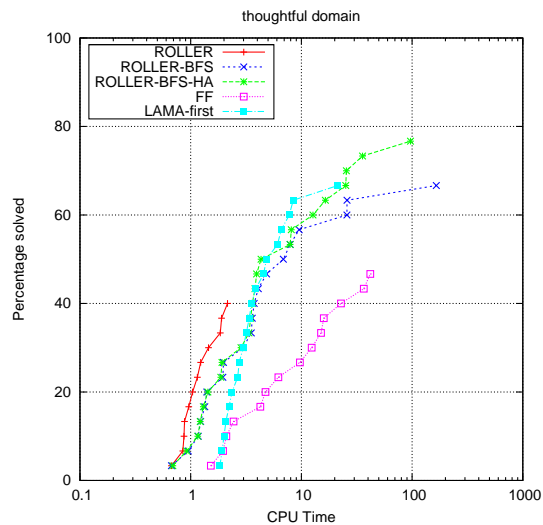

Figure 22: Percentage of solved problems when increasing time for evaluating the scalability performance in the *Thoughtful* domain.

### 4.2.14 TPP DETAILS

TPP stands for *Traveling Purchase Problem*, which is a generalization of the *Traveling Salesman Problem*. Tasks in the domain consist in selecting a subset of markets to satisfy the demand for a set of goods. The selection of markets should try to optimize the routing and the purchasing costs of the goods. In the STRIPS version, the graph that connects markets has equal costs for all arcs. Nevertheless, the domain is still interesting because it is difficult for planners to scale when increasing the number of goods, markets and trucks. The training set consists of thirty problems with a number of goods, trucks and depots varying from one to three and with load levels of five and six. The test set consists of the thirty problems used for planner evaluation at IPC-2006. The largest problem in this set has 20 goods, 8 trucks, 8 markets with a load level of six.

ROLLER, GR-HA and DF-HA solved the 30 problems in the test set, but ROLLER performs faster than the other two, achieving similar plan lengths. Besides, ROLLER outperforms the rest of the planners and it is two orders of magnitude faster than FF. The main reason is the overwhelming branching factor of the large problems together with the fact that FF heuristic falls into big plateaus in this domain. Greedy (depth-first) approaches perform better because they avoid the effect of these plateaus. Additionally, ROLLER achieved competitive quality scores and average plan length in the first solution and the anytime evaluation. ROLLER-BFS and ROLLER-BFS-HA got very bad results for this domain because of the imprecision of the `FF` heuristic. Figure 23 shows the percentage of solved problems while increasing the CPU time.

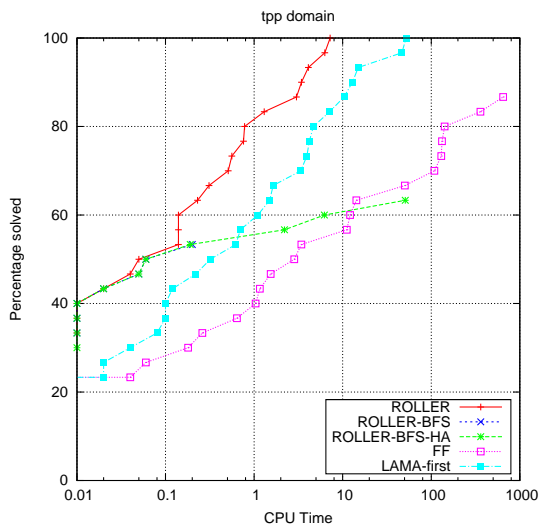

Figure 23: Percentage of solved problems when increasing time for evaluating the scalability performance in the *TPP* domain.

The learned DCK is compact and useful for reducing the number of evaluations, as shown by `ROLLER` performance. For instance, the `DRIVE` binding tree recognizes perfectly when a truck in market A does not need to go to a market B because there is already a truck in B handling the goods of that market. In these situations, the state with the truck in B has a helpful action `DRIVE`, meaning the truck B has something to deliver.

## 5. Lessons Learned from the IPC

IPC-2008 included a specific track for planning systems that benefit from learning. Thirteen systems took part in this track including a previous version of `ROLLER` (De la Rosa et al., 2009) that achieved the 7th position. This version was an upgrade of the original `ROLLER` system (De la Rosa et al., 2008). The first version proposed the *EHC-Sorted* algorithm as an alternative to the H-Context Policy, but it was not effective in many domains. The competing version tried to recommend ordering for applying actions from the relaxed plans. This idea, although initially appealing, was not a good choice because its usefulness strongly depends on the fact that the relaxed plan contains the right actions. After the competition we completed an analysis of the `ROLLER` performance to diagnose and strengthen its weak points. The system resulting from these improvements is the `ROLLER` version described in this article. One example of the `ROLLER` improvements is the results obtained at the *Thoughtful* and *Matching Blocksworld* domains. At IPC-2008, `ROLLER` failed to solve all the problems from the *Thoughtful* domain and it only solved two problems from the *Matching Blocksworld*. As reported in section 4, the current version of `ROLLER` solves 23 and 19 problems respectively in these domains. In addition, the current version of `ROLLER` outperforms `LAMA` and `FF` in the *Park-* *ing* domain by one order of magnitude. The improvements of ROLLER overcome limitations of the version submitted to IPC-2008 in three aspects:

- Robustness to wrong DCK. Issues discussed in Section 2 are all decisions that introduce biases in the learning process making learning of DCK a complex task. In fact, no competitor at IPC-2008 was able to learn useful DCK for all the domains. Furthermore, in many domains the learned DCK damaged the performance of the baseline planner. This was the case of ROLLER. As we described in the paper, we have strengthened ROLLER against wrong DCK by proposing two versions of a modified BFS algorithm that combine the learned DCK with a numerical heuristic. The combination of DCK and the heuristic makes the planning process more robust to imperfect and/or incorrectly learned knowledge. A similar approach was followed by the winner of the best learner award, OBTUSE WEDGE (Yoon et al., 2008).

- Efficiency of the baseline. The overall competition winner was PBP (Gerevini, Saetti, & Vallati, 2009) a portfolio of state-of-the-art planners that learns which planner and settings are the best ones for a given planning domain. As a result, the performance of this competitor was never worse than the performance of a state-of-the-art planner. At IPC-2008 the baseline performance of ROLLER was far from being competitive with state-of-the art planners because ROLLER algorithms were coded in LISP. To overcome this weakness we optimized the implementation of ROLLER using C code that outperformed our IPC-2008 results in all domains.

- Definition of significant training sets. Training examples are extracted from the experience collected while solving problems of a training set. Therefore, the quality of the training examples depends on the quality of the problems used for training. At IPC-2008 the training problems were fixed by the organizers and, in many domains, they were too large for the ROLLER system to extract useful DCK. In this paper we have created training problems using random generators to build useful training sets for the ROLLER system for each domain.

- Selection of training examples. Relational classifiers induce a set of rules/trees that model regularities in the training data. For the case of forward state-space search planning not all best-cost solutions for a problem may be used as training data, because this leads to alternatives that will confuse the learner. To avoid this, training data should be cleaned before being used by the learning algorithm. The ranking and solution selection proposed in this article is an option to give the learner training data with clearer regularities.

Additionally, ROLLER performed poorly in the *Sokoban* and *N-puzzle* domains. Traditionally, useful DCK for these domains has the form of numeric functions, such as the Manhattan distance, which provides a lower-bound for the solution length. In general, action policies are inaccurate in these domains, because they lack knowledge about the trajectory to the goals. Currently, we are still unable to learn useful DCK for ROLLER in these domains. A possible future direction is to introduce not only goals but subgoals (e.g. landmarks) in the *helpful context* with the aim of capturing some of this knowledge.

## 6. Related Work

Our approach is strongly inspired by the way Prodigy (Veloso et al., 1995) models DCK. In the Prodigy architecture, the action selection is a two-step process: first, Prodigy selects the uninstan-tiated operator to apply, and second, it selects the bindings for that operator. Both selections can be guided by DCK in the form of control rules (Leckie & Zukerman, 1998; Minton, 1990). We have returned to this idea of the two-step action selection because it allows us to define the learning of planning DCK as a standard classification task and therefore to solve this learning task with an off-the-shelf classification technique such as *relational decision trees*. Nevertheless, ROLLER does not need to distinguish among different kinds of nodes as Prodigy does, because ROLLER performs a standard forward heuristic search in the state space where all the search nodes are of the same type.

Relational decision trees have been previously used to learn action policies in the context of Relational Reinforcement Learning (RRL) (Dzeroski, De Raedt, & Blockeel, 1998). In comparison with the DCK learned by ROLLER, RRL action policies present two limitations when solving planning problems. First, in RRL the learned knowledge is targeted to a given set of goals, therefore RRL cannot directly generalize the learned knowledge for different goals within a given domain. Second, since training examples in RRL consist of explicit representations of the states, RRL needs to add extra background knowledge to learn effective policies in domains with recursive predicates such as *Blocksworld*.

Previous works on learning generalized policies (Martin & Geffner, 2004; Yoon et al., 2008) succeed in addressing these two limitations of RRL. First, they introduce planning goals in the training examples. In this way the learned policy applies for any set of goals in the domain. Second, they change the representation language of the DCK from *predicate logic* to *concept language*. This language makes capturing decisions related to recursive concepts easier. Alternatively, ROLLER captures effective DCK in domains like *Blocksworld* without varying the representation language. ROLLER implicitly encodes states in terms of the set of helpful actions of the state. As a result, ROLLER can benefit directly from off-the-shelf relational classifiers that work in predicate logic. This fact makes learning times shorter and the resulting policies easier to read.

Recently, other techniques have also been developed to improve the performance of heuristic planners:

- Learning Macro-actions (Botea, Enzenberger, Müller, & Schaeffer, 2005; Coles & Smith, 2007) are the combination of two or more operators that are considered as new domain operators in order to reduce the search tree depth. However, this benefit decreases with the number of new macro-actions added because they enlarge the branching factor of the search tree causing the *utility problem* (Minton, 1990). Other approaches overcome this problem, applying filters that decide on the applicability of the macro-actions (Newton, Levine, Fox, & Long, 2007). Two versions of this work participated in the learning track of IPC-2008, obtaining third and fourth place. One advantage of macro-actions is that the learned knowledge can be exploited by any planner. Thus, approaches which learn generalized policies could also benefit from macro-actions. Nevertheless, as far as we know, this combination has not been tried for improving heuristic planners.

- Learning domain-specific heuristic functions: In this approach (Yoon, Fern, & Givan, 2006; Xu, Fern, & Yoon, 2007), a state-generalized heuristic function is obtained from examples of solution plans. The main drawback of learning domain-specific heuristic functions is that the result of the learning algorithm is difficult to understand by humans which makes the verification of the learned knowledge difficult. On the other hand, the learned knowledge is easy to combine with existing domain-independent heuristics. A slightly different approach consists of learning a ranking function for greedy search algorithms (Xu, Fern, & Yoon, 2009, 2010). At each step of a greedy search, the current node is expanded and the child node with the highest rank is selected to be the current node. In this case, the ranking function is iteratively estimated in an attempt to cover a set of solution plans with the greedy algorithm.

- Learning task decomposition: This approach learns how to divide planning tasks of a given domain into smaller subtasks that are easier to solve. Techniques for reachability analysis and landmark extraction (Hoffmann, Porteous, & Sebastia, 2004) are able to compute intermediate states that must be reached before satisfying the goals. However, it is not clear how to systematically exploit this knowledge to build good problem decompositions. Vidal et al. (2010) consider this as an optimization problem and use a specialized optimization algorithm to discover good decompositions.

In general, any system that learns planning DCK has to deal with ambiguity in the training examples, because a given planning state may present many good actions. Trying to learn DCK that selects one action over other, inherently equal, is a complex learning problem. To cope with ambiguous training data ROLLER created a function that ranks solutions with the aim of learning from the same kind of solutions. A different approach is followed by Xu et al. (2010) who generate training examples from partially ordered plans.

## 7. Conclusions and Future Work

We have presented a new technique for reducing the number of node evaluations in heuristic planning based on learning and exploiting generalized policies. Our technique defines the process of learning generalized policies as a two-step classification and builds domain-specific relational decision trees that capture the action to be selected in the different planning contexts. In this work, planning contexts are specified by the helpful actions of the state, the pending goals and the static predicates of the problem. Finally, we have explained how to exploit the learned policies to solve classical planning problems, applying them directly or combining them with a domain independent heuristic in a lookahead strategy for the BFS algorithm. This work contributes to the state-of-the-art of learning-based planning in three ways:

1. Representation. We propose a new encoding for generalized policies that is able to capture efficient DCK using predicate logic. As opposed to previous works that represent generalized policies in predicate logic (Khardon, 1999), our representation does not need extra background knowledge (support predicates) to learn efficient policies for the *Blocksworld* domain. Besides, encoding states with the set of helpful actions is frequently more compact and furthermore, this set normally decreases when the search has fewer goals left. Thus, the process of matching DCK becomes faster when the search advances towards the goals.

2. Learning. We have defined the task of learning a generalized policy as a two-step standard classification task. Thus, we can learn the generalized policy with an off-the-shelf tool for building relational classifiers. Results in this paper are obtained with the TILDE system (Blockeel & De Raedt, 1998), but any other tool for learning relational classifiers could have been used. Because of this, advances in relational classification can be applied in a straightforward manner in ROLLER to learn faster and better planning DCK.

3. Planning. We have explained how to extract an action ordering from an *H-Context Policy* and we have shown how to use this ordering to reduce node evaluations: (1) in the algorithm *Depth-First H-Context Policy* that allows a direct application of the H-Context policies; and (2) in the *H-Context Policy Lookahead BFS*, which combines the policy with a domain-independent heuristic within a BFS algorithm. In addition, we have included a modified version of this algorithm (ROLLER-BFS-HA) that only considers helpful successors in order to reduce the number of evaluations in domains where helpful actions are good.

Experimental results show that our approach improved the scalability of the baseline heuristic planners FF and LAMA (winner of IPC-2008) over a variety of IPC domains. This effect is more evident in domains where the learned DCK presents good quality, e.g. *Blocksworld* and *Parking*. In these domains the direct application of the learned DCK saves large amounts of node evaluations achieving impressive scalability performance. Moreover, using the learned DCK in combination with a domain-independent heuristic in a BFS algorithm achieves good quality solutions. When the quality of the learned DCK is poor, planning with the direct application of the policy fails to solve many problems, mainly the largest ones which are more difficult to solve without a reasonable guide. Unfortunately, the only current mechanism for quantifying the quality of the learned DCK is evaluating it against a set of test problems. Therefore, a good compromise solution is combining the learned DCK with domain-independent heuristics.

In some domains, the DCK learned by ROLLER presents poor quality because the helpful context is not able to represent concepts that are necessary in order to discriminate between good and bad actions. This problem frequently arises when the arguments of the good action do not correspond to the problem goals or the static predicates. We plan to study refinements to our definition of the helpful context to achieve good DCK in such domains. One possible direction is extending the helpful context with subgoal information such as landmarks (Hoffmann et al., 2004) of the relaxed plan. Moreover, the use of decision trees introduces an important bias in the learning step. Algorithms for tree learning only insert a new query in the tree if doing so produces a significant information gain. However, in some domains this information gain can only be obtained by the conjunction of two or more queries. Finally, we are currently providing the learner with a fixed distribution of training examples. In the near future, we plan to explore how the learner can generate the most convenient distribution of training examples according to a target planning task as proposed by Fuentetaja and Borrajo (2006).

## Acknowledgments

This work has been partially supported by the Spanish MICIIN project TIN2008-06701-C03-03 and the regional CAM-UC3M project CCG08-UC3M/TIC-4141.

## Appendix A. DCK Usefulness Results

| | DEPH-FIRST | | | BEST-FIRST | | | HELPFUL BEST-FIRST | | |
|---|---|---|---|---|---|---|---|---|---|
| Domains | roller | gr-ha | df-ha | roller-bfs | lh-bfs | bfs | roller-bfs-ha | lh-bfs-ha | bfs-ha |
| Blocksworld (30) | **29.87** | 0.03 | 0.00 | 2.47 | 0.00 | 0.00 | 2.40 | 0.00 | 0.00 |
| Depots (22) | **19.42** | 7.70 | 3.61 | 10.51 | 5.32 | 2.45 | 10.98 | 5.27 | 7.08 |
| Gold-miner (30) | **28.00** | 0.00 | 0.00 | 0.04 | 0.05 | 0.00 | 7.35 | 0.00 | 0.00 |
| Matching-BW (30) | **20.88** | 0.00 | 0.00 | 3.82 | 0.98 | 3.90 | 3.72 | 2.20 | 5.81 |
| Parking (30) | **28.57** | 1.23 | 0.00 | 22.72 | 0.26 | 0.01 | 23.60 | 0.26 | 0.04 |
| Rovers (30) | **25.99** | 10.09 | 7.73 | 14.58 | 5.52 | 0.14 | 18.12 | 17.17 | 8.21 |
| Satellite (30) | **27.97** | 2.93 | 1.69 | 16.09 | 3.05 | 0.08 | 21.93 | 2.68 | 2.90 |
| Storage (30) | 11.02 | 8.03 | 8.08 | 11.53 | 10.56 | 8.97 | **16.12** | 7.00 | 7.00 |
| Thoughtful(30) | 11.91 | **13.15** | 0.00 | 11.30 | 8.90 | 2.31 | 11.89 | 9.05 | 3.04 |
| TPP (30) | **29.50** | 10.86 | 11.45 | 14.83 | 8.67 | 5.00 | 13.97 | 7.54 | 6.13 |
| Total | **233.13** | 54.02 | 32.56 | 107.89 | 43.31 | 22.86 | 130.08 | 51.17 | 40.21 |

Table 9: Problems solved for the DCK usefulness evaluation.

| | DEPH-FIRST | | | BEST-FIRST | | | HELPFUL BEST-FIRST | | |
|---|---|---|---|---|---|---|---|---|---|
| Domains | roller | gr-ha | df-ha | roller-bfs | lh-bfs | bfs | roller-bfs-ha | lh-bfs-ha | bfs-ha |
| Blocksworld (30) | **29.83** | 0.06 | 0.00 | 8.00 | 0.00 | 0.00 | 8.00 | 0.00 | 0.00 |
| Depots (22) | 8.82 | 8.21 | 3.02 | 12.68 | 14.84 | 12.37 | 13.20 | 16.06 | **19.97** |
| Gold-miner (30) | **19.78** | 0.00 | 0.00 | 11.50 | 17.00 | 15.41 | 16.67 | 0.00 | 0.00 |
| Matching-BW (30) | 11.53 | 0.00 | 0.00 | 12.76 | 6.35 | 13.84 | **17.21** | 9.39 | 16.54 |
| Parking (30) | 21.53 | 8.20 | 0.01 | **26.92** | 6.33 | 6.14 | **26.92** | 6.33 | 8.18 |
| Rovers (30) | 21.54 | 26.34 | 25.51 | 21.94 | 24.63 | 10.75 | 25.71 | 26.32 | **29.63** |
| Satellite (30) | **28.03** | 17.36 | 9.58 | 22.60 | 21.48 | 14.64 | 27.60 | 22.31 | 22.42 |
| Storage (30) | 13.43 | 8.02 | 8.00 | 15.59 | 16.87 | **19.23** | 15.66 | 8.59 | 9.31 |
| Thoughtful(30) | 6.89 | 12.40 | 0.00 | 16.56 | 13.35 | 10.60 | **19.48** | 14.91 | 11.75 |
| TPP (30) | 25.46 | **27.92** | 23.75 | 13.94 | 22.04 | 8.71 | 16.20 | 24.36 | 13.88 |
| Total | **186.84** | 108.51 | 69.87 | 162.49 | 142.89 | 111.69 | 186.65 | 128.27 | 132.35 |

Table 10: Quality scores for the DCK usefulness evaluation.